\documentclass{article} 
\usepackage{iclr2026_conference,times}

\usepackage{amsmath,amsfonts,bm}

\def\eqref#1{equation~\ref{#1}}

\def\1{\bm{1}}

\DeclareMathAlphabet{\mathsfit}{\encodingdefault}{\sfdefault}{m}{sl}
\SetMathAlphabet{\mathsfit}{bold}{\encodingdefault}{\sfdefault}{bx}{n}

\usepackage{hyperref}
\usepackage{url}

\usepackage{booktabs}
\usepackage{threeparttable}
\usepackage{graphicx}
\usepackage{caption}
\usepackage{multirow}
\usepackage{placeins}
\usepackage{makecell}

\title{FastAvatar: Towards Unified and Fast 3D Avatar Reconstruction with Large Gaussian Reconstruction Transformers}

\author{Yue Wu$^{1, 2}$ \quad
Xuanhong Chen$^{3}$\thanks{Corresponding authors} \quad
Yufan Wu$^{3}$ \quad
Wen Li$^{4}$ \quad
Yuxi Lu$^{1}$ \quad
Kairui Feng$^{1,2}$ 
\\
Tongji University$^{1}$ \quad 
Shanghai Innovation Institute$^{2}$ \quad 
Shanghai Jiao Tong University$^{3}$ \quad 
AKool$^{4}$\\
{\tt\small yuewu@tongji.edu.cn, chenxuanhong@sjtu.edu.cn, kelvinfkr@tongji.edu.cn}
}

\iclrfinalcopy 
\begin{document}

\maketitle
\begin{center}
    \centering
    \captionsetup{type=figure}
    \includegraphics[width=1\textwidth]{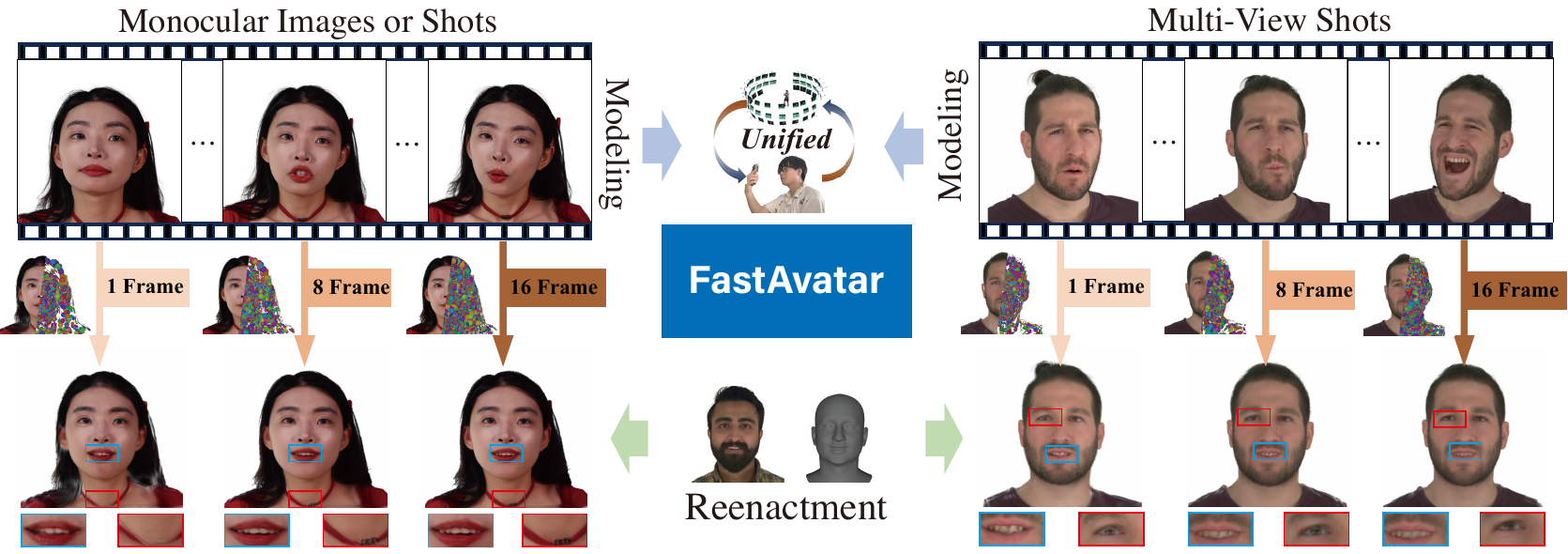}
    \captionof{figure}{Unlike existing 3D Avatar methods that can only process fixed-length data, FastAvatar achieves incremental reconstruction. 
  It can strike a good balance between modeling quality and inference speed based on available data volume, delivering high-quality models with sufficient data while providing viable reconstruction results at high speed even with limited data.
    }
    \label{teaser}
\end{center}

\begin{abstract}
Despite significant progress in 3D avatar reconstruction, it still faces challenges such as high time complexity, sensitivity to data quality, and low data utilization. 
We propose~\textbf{FastAvatar}, a feedforward 3D avatar framework capable of flexibly leveraging diverse daily recordings (e.g., a single image, multi-view observations, or monocular video) to reconstruct a high-quality 3D Gaussian Splatting (3DGS) model within seconds, using only a single unified model. 
The core of FastAvatar is a Large Gaussian Reconstruction Transformer (LGRT) featuring three key designs: 
First, a 3DGS transformer aggregating multi-frame cues while injecting initial 3D prompt to predict the corresponding registered canonical 3DGS representations; 
Second, multi-granular guidance encoding (camera pose, expression coefficient, head pose) mitigating animation-induced misalignment for variable-length inputs;
Third, incremental Gaussian aggregation via landmark tracking and sliced fusion losses. 
Integrating these features, FastAvatar enables incremental reconstruction, i.e., improving quality with more observations without wasting input data as in previous works.  
This yields a quality-speed-tunable paradigm for highly usable 3D avatar modeling. 
Extensive experiments show that FastAvatar has a higher quality and highly competitive speed compared to existing methods. Code and models are available at \url{https://github.com/TyrionWuYue/FastAvatar}.
\end{abstract}

\section{Introduction}
\label{sec:intro}

Creating photorealistic 3D avatar reconstruction is one of the fundamental problems in computer vision and graphics. Contemporary methods~\cite{avat3r, LAM,monogaussianavatar,DBLP:conf/cvpr/QianKS0GN24,DBLP:conf/nips/Pan0L0ZLSML24,DBLP:conf/cvpr/Wen0RSW24,DBLP:conf/cvpr/QianWM0024,DBLP:conf/cvpr/JiangSW0HZYX24,DBLP:conf/cvpr/Hu00ZLZN24,DBLP:journals/corr/abs-2503-10625} for 3D avatars have made significant advancements in 3D representation and modeling quality. 
However, these approaches commonly suffer from drawbacks such as data sensitivity (e.g., requiring richly expressive data), high time complexity, and low data utilization efficiency. 
These issues, pose severe challenges to the low-cost application of 3D avatars.

Three factors hinder existing 3D avatar methods from addressing the aforementioned challenges:
1)~\textit{Inability to Leverage Prior Knowledge}: Although contemporary 3D avatars have widely adopted efficient representations like 3DGS~\cite{3dgs}, they still primarily rely on per-scene optimization. 
This approach fails to utilize ``experience" from similar scenes, preventing the acquisition of good initial values to accelerate optimization. 
Furthermore, since all model information stems solely from the input observations, missing data cannot be reconstructed, resulting in a heavy dependence on complete 3D observations.
Daily captures, however, often contain significant information gaps. 
2)~\textit{Low Accuracy in Observation Alignment}: 3D avatar methods typically depend entirely on parametric proxy models (e.g., 3DMM/FLAME~\cite{DBLP:conf/siggraph/BlanzV99,flame}) for coarse view alignment. 
The precision of this alignment is critical for effective modeling; For instance, GaussianAvatars~\cite{DBLP:conf/cvpr/QianKS0GN24} even requires the proxy model to provide detailed meshes for hair. 
However, these parametric models are susceptible to limitations in representational capacity (e.g., blendshapes from 3D scan databases), lighting conditions, and data quality, often failing to produce highly accurate proxy 3D models. 
Using this proxy without refinement leads to poor robustness, hindering unified adaptation to diverse data sources (e.g., light fields, smartphones, DSLR cameras).
3)~\textit{Inadequate Handling of Variable-Length Data}:
Optimization-based 3D avatar methods typically require input data of a minimum specific length (typically at least 30 seconds at 25fps). Insufficient data often leads to modeling failure, resulting in severely limited capability to process few-shot data (e.g., 1 frame, 4 frames). 
Meanwhile, recently proposed feedforward-style methods~\cite{avat3r, LAM} are usually designed for fixed-length inputs for training convenience. 
For instance, LAM~\cite{LAM} can only process single-frame input, and Avat3r~\cite{avat3r} is fixed to handling exactly 4 frames. 
However, real-world data can consist of any arbitrary number of frames. 
The inability to process inputs of variable lengths will result in wasted valuable observation data, consequently limiting modeling quality.

To pursue data-efficient, high-quality, and fast 3D avatar reconstruction, we propose~\textit{FastAvatar}. 
It enables direct feedforward reconstruction of animatable avatars within seconds from arbitrary-length input frames and can incrementally leverage additional observation data.
The core of FastAvatar is a~\textbf{L}arge~\textbf{G}aussian~\textbf{R}econstruction~\textbf{T}ransformer (LGRT). 
It can align and aggregate variable-length facial inputs based on head pose and camera pose, then generate high-quality Gaussian model groups using coarse 3D positional prompts. 
Finally, these groups can be flexibly fused into a single 3DGS avatar model according to quality requirements and computational resources.
Notably, compared to LAM~\cite{LAM} and Avat3r~\cite{avat3r}, FastAvatar handles variable-length observation data with greater model flexibility and higher data utilization efficiency. 
Unlike VGGT~\cite{vggt}, FastAvatar is capable of directly generating 3DGS avatars and can achieve granular 3D model aggregation.
Benefiting from the successful architecture of VGGT, LGRT is designed as a variant of the VGGT structure. We replace the unstable Dense Prediction Transformer (DPT)~\cite{DBLP:conf/iccv/RanftlBK21} with an MLP that directly predicts canonical 3DGS models, while adopting 3D parametric model (e.g., FLAME) mesh vertices as positional prompts for the output. These improvements maximize adaptability for the prediction of the 3DGS avatar.
Instead of relying solely on single camera pose encoding, 3D avatar reconstruction demands higher requirements for input data alignment. Therefore, we additionally incorporate expression coefficients and head pose as positional encoding for input observations, enabling more precise cross-frame information aggregation.
Critically, we propose a landmark tracking loss and sliced fusion loss to efficiently supervise the model for enhanced aggregation accuracy while enabling incremental 3DGS models fusion.
Integrating these key designs, our model pioneers incremental 3D avatar reconstruction, meaning it can continuously ingest new observational data to progressively refine modeling quality.

Extensive experiments demonstrate that our model achieves highly competitive 3D reconstruction quality compared to state-of-the-art methods. It uniquely accomplishes incremental 3D avatar reconstruction, currently unattainable by existing approaches, and holds promise for delivering favorable solutions in the quality-speed trade-off paradigm.

\section{Related Work}

\subsection{3D-based Head Avatar Reconstruction}

FLAME-based~\cite{flame,feng2021learning,danvevcek2022emoca,ma2024cvthead,cudeiro2019capture} techniques utilize a parametric model in head reconstruction, allowing for effective expression control but struggle to represent details (e.g., eyes, teeth) and limited to single-view.
Since neural radiance fields have demonstrated strong ability to synthesis photo-realistic images, some method~\cite{zielonka2023instant, shao2023tensor4d, athar2023flame,muller2022instant} have adopted NeRF~\cite{mildenhall2021nerf} for head reconstruction, which perform higher fidelity, particularly in modeling fine-scale details like hair.
However, NeRF-based approaches~\cite{athar2022rignerf,guo2021ad,liu2022semantic} often suffer from a significant issue with head rendering speed limitations and extensive training data. Recently, 3DGS~\cite{3dgs} has demonstrated superior performance surpassing NeRF in both novel view synthesis quality and rendering speed.
Approaches ~\cite{DBLP:conf/cvpr/QianKS0GN24, monogaussianavatar, xu2024gaussian, wang2025gaussianhead, DBLP:conf/cvpr/WuCLJWFCLHZNZ25} generate photorealistic head avatars that allow full control over expressions and poses using multi-view videos. Another line of research places explicit emphasis on identity preservation~\cite{gerogiannis2025arc2avatar, zheng2025headgap, zielonka2025synthetic}.
Despite 3DGS’s impressive performance, it requires multi-frame data for identity-specific training and lacks flexibility, necessitating separate models for single-view and multi-view scenarios. In contrast, our FastAvatar achieves ultra-fast 3D head avatar reconstruction with a unified model.

\subsection{Feed-forward Reconstruction Model}
Traditionally, 3D reconstruction and view synthesis, depending on optimization-based approaches such as Structure-from-Motion~\cite{schonberger2016structure} and Multi-View Stereo~\cite{schonberger2016pixelwise}, are often computationally intensive, slow to converge, and reliant on precisely calibrated dataset, limiting their applications in real-world scenarios.
Recently, series of research~\cite{wang2024dust3r, chen2024mvsplat, liu2024meshformer,ye2024no,hong2023lrm,charatan2024pixelsplat,szymanowicz2024splatter,tang2024lgm,jin2024lvsm,zhang2024monst3r,DBLP:journals/corr/abs-2505-23716} initiate a new research paradigm termed Feed-forward 3D reconstruction model.
DUSt3R~\cite{wang2024dust3r} introduces a method for dense and unconstrained stereo 3D reconstruction, operating without prior camera calibration or viewpoint poses. 
VGGT~\cite{vggt} uses a large feed-forward transformer to effectively predict all key 3D attributes from a single image or multiple images.
While feed-forward networks excel in generic 3D reconstruction, their application to 3D head avatar reconstruction is still nascent and warrants  systematic exploration.
LAM~\cite{LAM} introduces a feed-forward framework that reconstructs an animatable gaussian head from a single image, allowing animation and rendering without additional post-processing.
Avat3r~\cite{avat3r} regresses animatable 3D head avatar from just a few input images, reducing compute requirements during inference.
A key challenge in Feed-forward Head Reconstruction Model is the absence of a unified framework to handle diverse real-world inputs, including monocular videos, sparse multi-view captures.
To address this gap, we propose a VGGT-style transformer architecture to jointly process different observation resource, achieving state-of-the-art performance.

\begin{figure*}[t]
    \centering
    \includegraphics[width=1.0\linewidth]{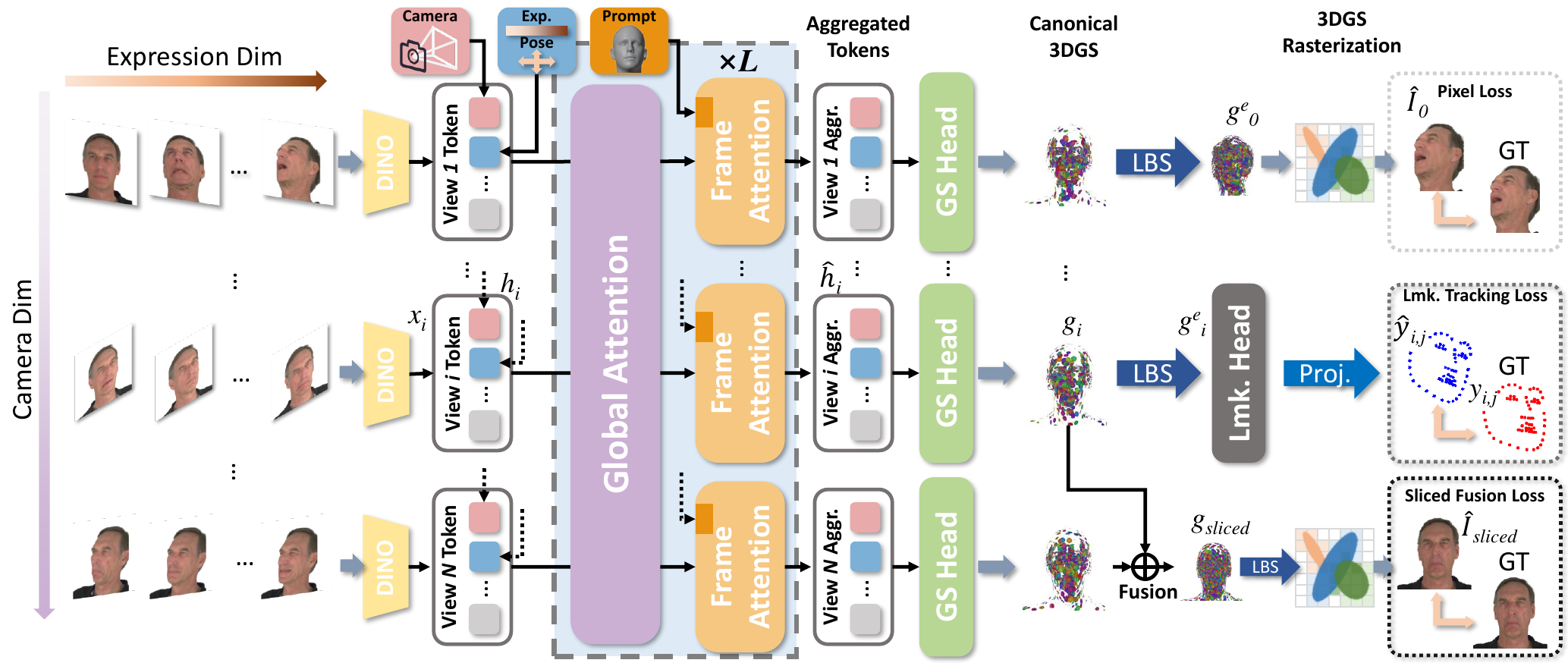}
    \caption{The core of FastAvatar is a Large Gaussian Reconstruction Transformer (LGRT), which can flexibly process input data with varying expressions, poses, and camera angles, aggregating them into a high-precision 3DGS avatar model. This capability is enabled by several key designs: the interleaving of global attention and frame attention to register complex input data while encoding 3D positional prompts; multi-granular positional information encoding; and the use of landmark tracking loss and sliced fusion loss, allowing the model to smoothly and incrementally fuse additional input data.
}
    \label{fig:framework}
\end{figure*}
\section{Methodology}

Daily observations are diverse and varied, such as single selfies, multi-angle selfies, video vlogs, etc. In summary, they are variable-length. Existing optimization-based 3D Avatar methods~\cite{avat3r,monogaussianavatar,DBLP:conf/cvpr/QianKS0GN24} cannot effectively handle overly short data (typically a single image). FastAvatar was specifically designed to address this scenario.

\subsection{Problem Definition and Notation}
FastAvatar $\mathcal{G}(\cdot)$ is a feed-forward avatar reconstruction framework designed to take any number of input observations and output a high-quality animatable 3DGS avatar:
\begin{equation}
    \mathcal{A} \leftarrow \mathcal{G}(I, \pi, z_{exp}, z_{pose}),
\end{equation}
where $(I_i)_{i=1}^{N}$ denotes an unordered sequence of $N$ RGB observations, with each $I_i\in{\mathbb{R}^{3\times H \times W}}$. 
$N$ does not exceed the maximum GPU capacity, typically $1\sim 16$ frames.
The corresponding facial expression and head pose are represented by $z_{exp}$ and $z_{pose}$, respectively. $\pi$ denotes the camera parameters and poses.
The output 3D Gaussian avatar representation $g$, including color $c\in\mathbb{R}^3$, opacity $o\in\mathbb{R}$, per-axis scale factors $s\in\mathbb{R}$, rotation $R\in\mathbb{R}^4$, importance score $m\in\mathbb{R}$, and points offest $O\in\mathbb{R}^3$, can be driven by any desired expression and pose from an arbitrary viewpoint using a differentiable rasterizer $\Psi$ \cite{3dgs}.

\subsection{Large Gaussian Reconstruction Transformer}
The core of FastAvatar is a Large Gaussian Reconstruction Transformer (LGRT). 
The LGRT is redesigned specifically for 3D avatar tasks, which demand finer granularity compared to SLAM-based environmental reconstruction. Moreover, the human subjects captured for 3D avatars cannot remain perfectly static and often exhibit rich dynamic characteristics (i.e., expressions, poses, etc.). The LGRT comprises 6 stages: facial feature extraction, face encoding, face aggregation and registration, 3DGS attribute generation, canonical 3DGS model fusion, and 3DGS rasterization.

\paragraph{Face Encoding}
FastAvatar encodes each face observation $I_i$ to a set of token $x_i$ through DINOv2~\cite{dinov2}. 
These tokens vary from head poses, facial expressions to camera poses, and undifferentiated aggregation would lead to over-smoothing and aliasing effects.
Therefore, FastAvatar introduces three critical encodings to label distinct facial tokens, facilitating subsequent aggregation. This process can be formulated as:
\begin{equation}
    h_i = \mathcal{U}\left(x_i, \mathrm{MLP}([\pi_i, z^{exp}_i, z^{pose}_i])\right),
\end{equation}
where $\mathcal{U}(\cdot)$ denotes concatenation along the dimensional axis.
$h_t$ denotes the encoded face tokens.
$\pi_i$, $z^{pose}_i$, and $z^{exp}_i$ represent the camera pose, head pose, and expression coefficients of $x_i$ respectively.
These are processed through a lightweight MLP layer for feature alignment and dimensionality alignment.
We obtain rough initial estimates of the three parameters through multi-view FLAME tracking.

\paragraph{Face Aggregation and Registration}
The core enabling component for constructing a dense 3D avatar from variable-length input data lies in the aggregation and registration of face tokens. 
The purpose of aggregation is to extract intra-token features while incorporating initialized 3D positional prompts. 
These positional prompts provide the LGRT with initial 3DGS positions, thereby accelerating 3D reconstruction. 
As illustrated in Figure~\ref{fig:framework}, aggregation is implemented through frame attention, composed of dual-stream DiT blocks~\cite{labs2025flux1kontextflowmatching,flux2024}, which aggregates intra-token information while fusing 3D positional prompts.
Face token registration serves as the fundamental operation for fusing multiple inputs. 
In Figure~\ref{fig:framework}, this is achieved via global attention, which aligns encoded face tokens to achieve 3D spatial registration and fusion. 
To enhance quality and accelerate convergence, we initialize our frame attention using weights from LAM's blocks.
Global attention and frame attention are interleaved in a cascaded architecture, with a total of $L$ pairs employed to process face tokens, ultimately yielding tokens suitable for generating the 3DGS representation $\{\hat{h}_0, \cdots, \hat{h}_N\}$.

\paragraph{Canonical 3DGS Model Fusion}
Following the aggregation and registration through global attention and frame attention, we obtain processed tokens $\hat{h}_i$ corresponding to each frame. These features are then fed into a GS Head (i.e., a two-layer MLP with shared weights across tokens) to predict the target 3DGS attributes $g_i$. 
The point cloud $g^e_i$ derived by driving $g_i$ through Linear Blend Skinning (LBS) expression deformation is then rendered via Gaussian splatting to obtain the reconstructed face $\hat{I}_i$.
Our approach extends beyond $g_i$; we further aggregate all ${g_0,\cdots,g_N}$:

\begin{equation}
    g_f = \mathcal{U}(g_1, g_2, \cdots, g_N).
\end{equation}
The fused $g_f$ integrates unique information from all perspectives (e.g., multi-view observations, diverse expressions), achieving optimal reconstruction quality. However, naive fusion would cause point cloud misalignment, ghosting artifacts, and color discrepancies. To address this, we introduce Landmark Tracking Loss and Sliced Fusion Loss to explicitly encourage proper alignment of Gaussian point clouds during aggregation and registration stages.

\paragraph{3DGS Pruning}
Although the 3DGS method achieves high-quality and real-time rendering, it often suffers from redundant memory consumption due to its explicit structure and tends to be more prone to overfitting because of the lack of smoothness bias in the neural network. This is especially problematic in our incremental reconstruction scenarios, where the number of GS points increases linearly with the number of input frames, leading to inefficient resource usage and limiting rendering speed. To address this, inspired by LP-3DGS~\cite{LP3DGS} and MaskGaussian~\cite{MaskGaussian} we apply Gumbel-Softmax~\cite{DBLP:conf/iclr/JangGP17} to sample one differentiable category, denoted as $\mathcal{M}_i\in\{0,1\}$. Then we integrate masks directly within the rasterization framework, effectively decoupling Gaussian presence from attributes such as opacity and shape. This mechanism prunes over 50\% of the GS primitives without degrading reconstruction quality, further improving rendering efficiency. To prune redundant 3D Gaussian primitives, we apply an L1 regularization term to the trainable mask, encouraging it to become sparse, formulated as $\mathcal{L}_{mask}=\frac{1}{N}\sum_{i=1}^N\lvert m \rvert$.

\subsection{Training Strategy}
\paragraph{Sliced Fusion Loss} To enable the model to take advantage of the richer information provided by multiple inputs, we introduce \textit{Sliced Fusion Loss}, allowing $\mathcal{G}$ to handle arbitrary numbers of input frames. 
Specifically, during training, we randomly sample one frame from the input to obtain a single frame-wise Gaussian representation $g_i$. 
In parallel, we randomly select $N_\text{sliced}$ frames from the input, where $N_\text{sliced}$ is less than the total number of input frames for memory efficiency, and fuse them to construct a multi-frame Gaussian representation $g_\text{sliced}$. 
Both $g_i$ and $g_\text{sliced}$ are rendered into RGB images using the camera parameters, expression coefficients, and head poses of all input and target frames, and the corresponding losses are computed.

\begin{equation}
    \hat{I}_i = \Psi\left(g_i, \pi_i, z^{\mathrm{exp}}_i, z^{\mathrm{pose}}_i\right),
\end{equation}
\begin{equation}
    \hat{I}_\text{sliced} = \Psi\left(g_\text{sliced}, \pi_i, z^{\mathrm{exp}}_i, z^{\mathrm{pose}}_i\right).
\end{equation}
The overall loss function consists of two components: one supervises the rendering quality of the constructed 3D Gaussian head, and the other supervises the combination of frame-wise Gaussian representations to ensure consistency across frames.

\paragraph{Pixel Losses} The rendered RGB images are supervised using photometric losses against the corresponding ground truth target images:
\begin{equation}
    \mathcal{L}_{rgb}=\left\Vert\hat{I}_i, I^{gt}\right\Vert_1 + \left\Vert\hat{I}_\text{sliced}, I^{gt}\right\Vert_1,
\end{equation}
\begin{equation}
    \mathcal{L}_{ssim} = \mathrm{SSIM}(\hat{I}_i, I^{gt}) + \mathrm{SSIM}(\hat{I}_\text{sliced}, I^{gt}),
\end{equation}
We also incorporate perceptual losses to encourage the emergence of more high-frequency details:
\begin{equation}
    \mathcal{L}_{lpips} = \mathrm{LPIPS}(\hat{I}_i, I^{gt}) + \mathrm{LPIPS}(\hat{I}_\text{sliced}, I^{gt}).
\end{equation}

\paragraph{Landmark Tracking Loss} Unlike novel view synthesis, canonical space registration is supervised directly on the input frames. The landmark tracking loss is introduced to encourage precise localization of facial landmarks throughout the input images:
\begin{equation}
    \mathcal{L}_{track}=\sum_{j=1}^M\sum_{i=1}^N\left\Vert y_{j,i}-\hat{y}_{j,i}\right\Vert.
\end{equation}


Our total loss is defined as follows:
\begin{equation}
    \mathcal{L} = \lambda_{1}\mathcal{L}_{rgb} + \lambda_{2}\mathcal{L}_{ssim} + \lambda_{3}\mathcal{L}_{lpips} + 
    \lambda_{4}\mathcal{L}_{track} + 
    \lambda_{5}\mathcal{L}_{mask},
\end{equation}
with $\lambda_{1}=0.8$, $\lambda_{2}=0.1$, $\lambda_{3}=0.1$, $\lambda_{4}=0.1$ and $\lambda_{5}=0.0005$.

\begin {table*}[t]
\centering
\small
\setlength{\tabcolsep}{0.8mm}
    \begin{tabular}{lcccccc}
    \toprule[1.5pt]
        {\textbf{Method}}  & \multicolumn{6}{c}{ 1 View } \\
        & PSNR$\hspace{0.2em}\uparrow$ & SSIM$\hspace{0.2em}\uparrow$ & LPIPS$\hspace{0.2em}\downarrow$ & Identity$\hspace{0.2em}\downarrow$ & FPS$\hspace{0.2em}\uparrow$ & Modeling Time (s)$\hspace{0.2em}\downarrow$\\

        \midrule
        LAM & \underline{17.30} & \underline{0.773} & \underline{0.149} & \underline{0.135} & \underline{125} & \textbf{0.31} \\
        
        MonoGaussianAvata & 11.83 & 0.631 & 0.620 & 0.432 & \textless 10 & \textgreater 100 \\

        GaussianAvatars & 16.35 & 0.740 & 0.332 & 0.299 & \textless 10 & \textgreater 100 \\

        FastAvatar & \textbf{20.08} & \textbf{0.860} & \textbf{0.143} & \textbf{0.116} & \textbf{240.17} & \underline{1.33} \\

        \midrule{\textbf{Method}} & \multicolumn{6}{c}{ 4 Views } \\
        & PSNR$\hspace{0.2em}\uparrow$ & SSIM$\hspace{0.2em}\uparrow$ & LPIPS$\hspace{0.2em}\downarrow$ & Identity$\hspace{0.2em}\downarrow$ & FPS$\hspace{0.2em}\uparrow$ & Modeling Time (s)$\hspace{0.2em}\downarrow$ \\
        \midrule

        LAM* & 16.69 & 0.743 & \underline{0.204} & \underline{0.167} & \underline{45} & \textbf{0.39} \\

        MonoGaussianAvatar & 12.71 & 0.798 & 0.437 & 0.368 & \textless 10 & \textgreater 100 \\

        GaussianAvatars & \underline{17.52} & \underline{0.802} & 0.340 & 0.278 & \textless 10 & \textgreater 100 \\

        FastAvatar & \textbf{22.12} & \textbf{0.880} & \textbf{0.094} & \textbf{0.098} & \textbf{101.62} & \underline{4.22} \\

        \midrule{\textbf{Method}} & \multicolumn{6}{c}{ 8 Views } \\
        & PSNR$\hspace{0.2em}\uparrow$ & SSIM$\hspace{0.2em}\uparrow$ & LPIPS$\hspace{0.2em}\downarrow$ & Identity$\hspace{0.2em}\downarrow$ & FPS$\hspace{0.2em}\uparrow$ & Modeling Time (s)$\hspace{0.2em}\downarrow$ \\
        \midrule

        LAM* & 16.59 & 0.718 & \underline{0.235} & \underline{0.206} &\underline{24} & \textbf{0.43} \\

        MonoGaussianAvatar & 13.11 & 0.650 & 0.493 & 0.298 &\textless 10 & \textgreater 100 \\

        GaussianAvatars & \underline{20.35} & \underline{0.820} & 0.320 & 0.252 & \textless 10 & \textgreater 100 \\

        FastAvatar & \textbf{22.19} & \textbf{0.880} & \textbf{0.093} & \textbf{0.097} & \textbf{52.28} & \underline{8.56} \\

        \midrule{\textbf{Method}} & \multicolumn{6}{c}{ 16 Views } \\
        & PSNR$\hspace{0.2em}\uparrow$ & SSIM$\hspace{0.2em}\uparrow$ & LPIPS$\hspace{0.2em}\downarrow$ & 
        Identity$\hspace{0.2em}\downarrow$ & FPS$\hspace{0.2em}\uparrow$ & Modeling Time (s)$\hspace{0.2em}\downarrow$ \\
        \midrule

        LAM* & 16.49 & 0.697 & \underline{0.265} & 0.238 & \underline{13} & \textbf{0.69} \\

        MonoGaussianAvatar & 15.81 & 0.721 & 0.406 & 0.202 & \textless 10 & \textgreater 100 \\

        GaussianAvatars & \underline{21.48} & \underline{0.873} & 0.281 & \underline{0.185} & \textless 10 & \textgreater 100 \\

        FastAvatar & \textbf{22.29} & \textbf{0.881} & \textbf{0.092} & \textbf{0.095} & \textbf{17.65} & \underline{26.06} \\

    \bottomrule[1.5pt]
    \end{tabular}
    \caption{The quantitative comparison among FastAvatar, LAM~\cite{LAM}, MonoGaussianAvatar~\cite{monogaussianavatar}, and GaussianAvatars~\cite{DBLP:conf/cvpr/QianKS0GN24} includes 3 critical metrics: Reconstruction quality (PSNR, SSIM, LPIPS); Modeling time: Duration required to reconstruct the 3DGS model; Inference speed: Animation rendering FPS of the output 3DGS model.}
\label{tab:main_comparison}
\end{table*}

\section{Experiments}

\subsection{Training}
We train FastAvatar on a multi-task dataset derived from NeRSemble \cite{nersemble}, which contains multi-person, multi-camera videos with a wide range of facial expressions. To encourage adaptability to diverse input settings, the dataset includes both monocular and multi-view subsets. Input-output pairs are constructed by sampling 16 frames each, either from a single video or from 12 camera views of the same subject. For each pair, a random subset of 1 to 16 input frames is further selected, enabling the model to robustly handle scenarios with sparse or varying numbers of input frames—such as real-world recordings with incomplete or irregular camera captures.

\subsection{Experimental Setups}

\paragraph{Task.}
We evaluate the model’s ability to generate a 3D head avatar for unseen subjects from various types of input, including a single image, an unordered and arbitrary number of monocular video frames, and multi-view frames.
\paragraph{Metrics.}
We employ three paired-image metrics to measure the quality of individual rendered images: Peak Signal-to-Noise (PSNR), Structural Similarity Index (SSIM), and Learned Perceptual Image Patch Similarity (LPIPS). We also evaluate identity preservation by computing similarity using ArcFace~\cite{deng2019arcface, DBLP:conf/mm/ChenCNG20, DBLP:journals/pami/ChenNLLZW24} features.
\paragraph{Baselines.} We compare FastAvatar with recent state-of-the-art systems for 3D head avatar generation across various tasks, including reconstruction from a single image, monocular video frames and multi-view frames.
LAM~\cite{LAM} A model that generates one-shot animatable Gaussian heads using a canonical Transformer with point-cloud representation and multi-scale cross-attention, enabling real-time, expression-consistent avatar animation and editing from a single image.
Avat3r~\cite{avat3r} A system for regressing animatable 3D head avatars from limited multi-view images by combining large reconstruction and foundation models with cross-attention layers to effectively model 3D facial dynamics and generalize across diverse data.
MonoGaussianAvatar~\cite{monogaussianavatar} and GaussianAvatar~\cite{DBLP:conf/cvpr/QianKS0GN24} are two representative optimization-based 3D Avatar methods, both of which use FLAME as a proxy 3D model, similar to ours. 

We conduct all comparison experiments on the same 48GB Nvidia RTX 4090 GPU. 
To ensure a fair comparison, we slightly modify the LAM renderer for single-shot input, enabling it to fuse information from multiple frames like FastAvatar, while keeping all other components consistent with the official repository. We retain the original model weights and confirm that its performance faithfully reflects the official version. We refer to this variant as LAM* in the following. Notably, with only 1 input frame, LAM* automatically reverts to the official LAM, which is designed for single-frame inputs, ensuring fair comparison.

\begin{figure*}[t]
    \centering
\includegraphics[width=1.0\linewidth]{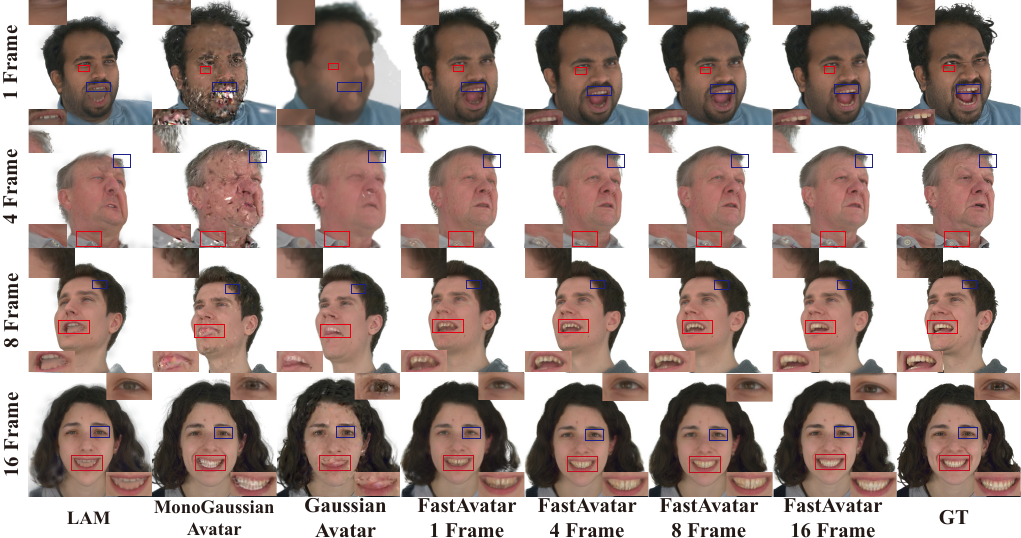}
  \caption{We benchmark FastAvatar against representative optimization-based methods (MonoGaussian Avatar~\cite{monogaussianavatar}, GaussianAvatar~\cite{DBLP:conf/cvpr/QianKS0GN24}) and feedforward approaches (LAM~\cite{LAM}). 
  Our results demonstrate the performance evolution across methods as the number of input views (referring to input images number) increases. Please zoom in for a better view.}
  \label{main_comparison}
\end{figure*}

\subsection{Qualitative Comparison}
Comparative results against LAM, MonoGaussian, and GaussianAvatar are presented in Figure~\ref{main_comparison}. 
We evaluate 4 distinct input configurations (1, 4, 8, and 16 views) by progressively increasing the number of input frames. 
Key observations indicate that while LAM yields performance roughly on par under single-view conditions, it fails to benefit from additional input views due to the lack of registration.
Conversely, optimization-based methods exhibit significant performance degradation with sparse inputs (e.g., 1 view), though their reconstruction quality improves progressively as more views become available. 
FastAvatar consistently outperforms the baseline across all view settings (1$\sim$16 views), while further enhancing its ability to capture fine-grained details—such as teeth gaps, wrinkles, and acne—as the number of views increases.

\subsection{Quantitative Comparison}
Table~\ref{tab:main_comparison} presents a quantitative comparison of the four methods. 
All approaches demonstrate substantial improvements in reconstruction metrics with increasing input frames, with both MonoGaussianAvatar and GaussianAvatar exhibiting gains in both subjective assessments (i.e., LPIPS) and objective metrics (i.e., PSNR, SSIM). 
This reaffirms the critical importance of richer input data for high-fidelity reconstruction. 
However, LAM shows an inverse relationship: 
as its input views increase, quantitative performance degrades, which can also be illustrated in Figure~\ref{fig:incremental_rec}.
Although LAM achieves impressive visual quality with single image (LPIPS: 0.149), its generative bias introduces pose and expression artifacts that compromise objective measurements. 
FastAvatar achieves highly competitive growth across both subjective and objective dimensions, validating our core hypothesis. 
Through architectural and training innovations, FastAvatar establishes an optimal equilibrium between generative capability (hallucinating plausible details under sparse inputs) and reconstruction fidelity (strict adherence to observed data given sufficient views).

\subsection{Incremental Reconstruction}
FastAvatar enables incremental improvement by accepting input sequences of any length and order. As more observations are added, the reconstruction quality continues to improve. In contrast, existing methods often require a fixed number of input frames, which reduces flexibility and may result in data loss. FastAvatar’s incremental design thus ensures both versatility and efficient data usage.

\begin{figure}[t]
  \centering
  \includegraphics[width=1.0\linewidth]{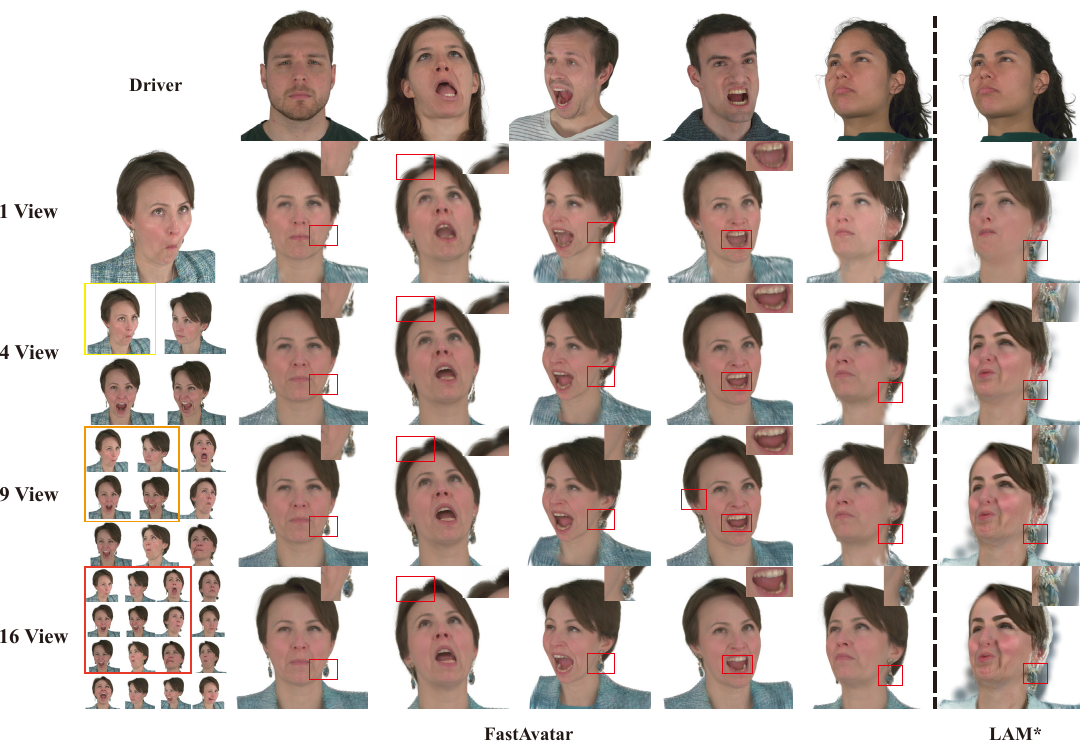}
  \caption{Reconstruction quality as the number of input observations increases. More observations improve reconstruction quality.}
  \label{fig:incremental_rec}
\end{figure}

As illustrated in Figure~\ref{fig:incremental_rec}, our method achieves superior expressiveness and rendering fidelity in avatar generation compared to the baselines, and demonstrates robust performance even for subjects wearing accessories.

Moreover, increasing the number of input views further improves the reconstruction of fine-grained details, such as hair and teeth gaps.
This incremental reconstruction allows the model to overcome the limitations of restricted viewpoints by leveraging more informative input frames—a capability that is difficult to achieve for models with fixed input forms. For example, in Figure~\ref{fig:incremental_rec}, the character's left-ear earring is not sufficiently observed with a small number of input views, but it is reliably reconstructed as the number of views increases.

\begin{figure*}[tbp]
    \centering
\includegraphics[width=1.0\linewidth]{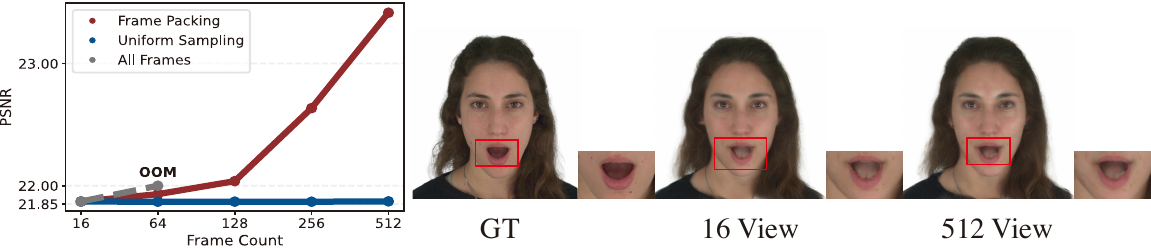}
  \caption{Performance on longer input sequences. Starting from strong reconstructions using only the 16 sparse input frames, incorporating the compressed additional frames further enhances fine-grained details (e.g., the oral cavity, which is absent in most frames). While uniform sampling fails to achieve this improvement, feeding all frames leads to OOM.}
  \label{long_input}
\end{figure*}

Unlike the sparse and randomly sampled data used in our main experiments, real-world sequences are highly continuous. Processing all frames with Global Attention imposes prohibitive computational and memory costs. A naive solution is to uniformly sample 16 frames, but this risks missing important information present in the remaining frames. Inspired by FramePack~\cite{zhang2025framepack}, we retain the 16 uniformly sampled frames as sparse inputs and compress all remaining frames into an aggregated token representation, which is then treated as an additional input. This preserves complementary details while keeping computation tractable, enabling FastAvatar to process hundreds of frames in a single feed-forward pass. Figure~\ref{long_input} shows that adding the compressed frames improves fine-grained details.

\subsection{Multi-view Observations Reconstruction}
\begin{figure}[tbp]
  \centering
\includegraphics[width=1.0\linewidth]{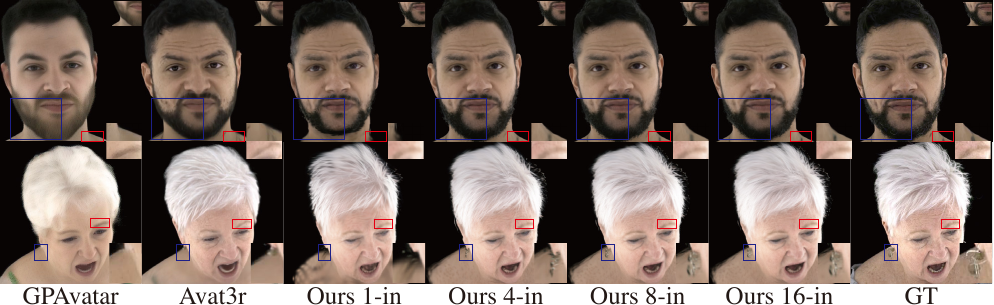}
  \caption{Visual comparison with Avat3r and GPAvatar~\cite{DBLP:conf/iclr/ChuLZYLLH24}. Please zoom in for a clearer view.}
  \label{fig:comparsion_avat3r}
\end{figure}

To further evaluate FastAvatar’s performance on the Multi-view Observations Reconstruction task, we create multi-view few-shot 3D head avatars for subjects from the Ava256~\cite{ava256} dataset, which was not used during training. 
We compare our results with those of the state-of-the-art method Avat3r~\cite{avat3r} and use the results provided in its original paper to ensure a fair comparison (since its implementation is not publicly available, we ensure fairness by directly using results from the Avat3r paper). 
The qualitative results are shown in Figure~\ref{fig:comparsion_avat3r}. 
Note that we only used images from Ava256 for tracking and to obtain FLAME parameters, without utilizing the provided informed encodings.  
Nevertheless, FastAvatar still achieves highly competitive results and benefits from additional multi-view inputs, producing more detailed reconstructions.
Close inspection reveals that Avat3r fails to preserve facial identity accurately, reconstructing consistently wider facial structures than observed in source inputs.

\begin{table}[htb]
\centering
\small
\setlength{\tabcolsep}{1mm}

\resizebox{\textwidth}{!}{%
    \begin{tabular}{lcccccc|cccccc}
    \toprule[1.pt]
        {\textbf{Method}}  & \multicolumn{6}{c}{ 1 View } & \multicolumn{6}{c}{ 4 View } \\
        & L1$\hspace{0.2em}\downarrow$ & 
        PSNR$\hspace{0.2em}\uparrow$ & SSIM$\hspace{0.2em}\uparrow$ & LPIPS$\hspace{0.2em}\downarrow$ & Identity$\hspace{0.2em}\downarrow$ & 
        \# GS (K)$\hspace{0.2em}\downarrow$ & L1$\hspace{0.2em}\downarrow$ &
        PSNR$\hspace{0.2em}\uparrow$ & SSIM$\hspace{0.2em}\uparrow$ & LPIPS$\hspace{0.2em}\downarrow$ & Identity$\hspace{0.2em}\downarrow$ & 
        \# GS (K)$\hspace{0.2em}\downarrow$ \\

        \midrule
        w/o sliced fusion loss & \underline{0.0373} & \underline{20.47} & \underline{0.861} & \textbf{0.123} & 0.158 & 20.7 & 0.0345 & \underline{21.69} & 0.857 & 0.131 & 0.138 & 82.8 \\
        
        w/o tracking loss & \textbf{0.0372} & \textbf{20.93} & \textbf{0.863} & 0.140 & 0.172 & 13.0 & \underline{0.0322} & 21.64 & 0.866 & 0.120 & \underline{0.128} & 41.2 \\

        w/o global attention & 0.0922 & 15.40 & 0.760 & 0.238 & 0.400 & \textbf{10.3} & 0.0462 & 19.49 & 0.828 & 0.162 & 0.270 & 40.1 \\

        w/o GS fusion & 0.0379 & 20.31 & 0.857 & \underline{0.136} & \textbf{0.138} & \underline{12.8} & 0.0467 & 18.94 & 0.838 & 0.157 & 0.185 & \textbf{12.5} \\

        w/o GS pruning & 0.0380 & 20.32 & 0.857 & 0.137 & \underline{0.144} & 21.7 & \underline{0.0322} & 21.67 & \underline{0.868} & \underline{0.112} & 0.130 & 86.7 \\

        Ours full & 0.0379 & 20.31 & 0.857 & \underline{0.136} & 0.148 & \underline{12.8} & \textbf{0.0303} & \textbf{21.86} & \textbf{0.871} & \textbf{0.107} & \textbf{0.118} & \underline{42.8} \\

        \midrule{\textbf{Method}} & \multicolumn{6}{c}{ 8 View } & \multicolumn{6}{c}{ 16 View } \\

        & L1$\hspace{0.2em}\downarrow$ & 
        PSNR$\hspace{0.2em}\uparrow$ & SSIM$\hspace{0.2em}\uparrow$ & LPIPS$\hspace{0.2em}\downarrow$ & Identity$\hspace{0.2em}\downarrow$ & 
        \# GS (K)$\hspace{0.2em}\downarrow$ & L1$\hspace{0.2em}\downarrow$ &
        PSNR$\hspace{0.2em}\uparrow$ & SSIM$\hspace{0.2em}\uparrow$ & LPIPS$\hspace{0.2em}\downarrow$ & Identity$\hspace{0.2em}\downarrow$ & 
        \# GS (K)$\hspace{0.2em}\downarrow$ \\
        \midrule

        w/o sliced fusion loss & 0.0386 & 21.12 & 0.849 & 0.144 & 0.151 & 165.4 & 0.0417 & 20.62 & 0.839 & 0.159 & 0.180 & 330.5 \\
        
        w/o tracking loss & \underline{0.0320} & 21.61 & 0.867 & 0.119 & \underline{0.124} & 78.6 & \underline{0.0322} & \underline{21.66} & 0.865 & 0.123 & \underline{0.129} & 164.2 \\

        w/o global attention & 0.0413 & 19.97 & 0.835 & 0.156 & 0.223 & 78.0 & 0.0405 & 20.06 & 0.830 & 0.167 & 0.210 & 155.7 \\

        w/o GS fusion & 0.0574 & 17.44 & 0.823 & 0.179 & 0.227 & \textbf{12.5} & 0.0682 & 16.25 & 0.811 & 0.196 & 0.259 & \textbf{12.4} \\

        w/o GS pruning & 0.0326 & \underline{21.63} & \underline{0.868} & \underline{0.110} & 0.130 & 173.4 & 0.0327 & 21.61 & \underline{0.867} & \underline{0.110} & 0.136 & 346.8 \\

        Ours full & \textbf{0.0297} & \textbf{21.95} & \textbf{0.871} & \textbf{0.103} & \textbf{0.118} & \underline{77.0} & \textbf{0.0293} & \textbf{22.04} & \textbf{0.876} & \textbf{0.102} & \textbf{0.118} & \underline{138.9} \\
        
    \bottomrule[1.pt]
    \end{tabular}
    }
    
    \caption{Ablation studies on key components of FastAvatar. The appendix visualizations are strongly recommended for better understanding.}
\label{tab:ablation-indices}
\end{table}
\subsection{Ablation Study}

To validate the effectiveness of each component in our method, we conduct both quantitative and qualitative ablation studies. The qualitative visualizations are provided in the appendix for space considerations. As shown in Table~\ref{tab:ablation-indices}, Global Attention is crucial for coordinating inter-frame information, while GS Fusion aggregates the GS points from each frame into a unified representation. Sliced Fusion Loss and Tracking Loss supervise GS registration, enforcing structural consistency and temporal coherence. Without these components, newly introduced frames fail to provide reliable information, leading to degraded registration and blurred outputs as the number of frames increases. Meanwhile, Gaussian Pruning removes redundant primitives, slightly improving performance while substantially accelerating rendering. Together, these mechanisms ensure effective inter-frame coordination, accurate registration, and efficient rendering.
\section{Conclusion}
In this paper, we present FastAvatar, a feed-forward 3D avatar reconstruction framework capable of constructing a high-quality animatable 3DGS avatar within seconds. Distinct from existing approaches, FastAvatar demonstrates a unique capability for incremental avatar reconstruction – flexibly leveraging incoming observations to progressively enhance reconstruction quality. We contend this represents a promising research trajectory. Three pivotal innovations enable this functionality:
Alternating Attention, augmented with fine-grained expression and pose encodings, achieves high-precision registration of unordered data;
The proposed Landmark Tracking Loss and Sliced Fusion Loss facilitate robust fusion of multiple 3DGS representations for superior modeling fidelity.
Experimental validation confirms FastAvatar's potential in these dimensions. 

\bibliography{iclr2026_conference}

@inproceedings{LAM,
  title={LAM: Large Avatar Model for One-shot Animatable Gaussian Head},
  author={
    Yisheng He and Xiaodong Gu and Xiaodan Ye and Chao Xu and Zhengyi Zhao and Yuan Dong and Weihao Yuan and Zilong Dong and Liefeng Bo
  },
  booktitle={SIGGRAPH},
  year={2025}
}

@inproceedings{vggt,
  title={VGGT: Visual Geometry Grounded Transformer},
  author={Wang, Jianyuan and Chen, Minghao and Karaev, Nikita and Vedaldi, Andrea and Rupprecht, Christian and Novotny, David},
  booktitle={Proceedings of the IEEE/CVF Conference on Computer Vision and Pattern Recognition},
  year={2025}
}

@article{3dgs,
  title={3d gaussian splatting for real-time radiance field rendering.},
  author={Kerbl, Bernhard and Kopanas, Georgios and Leimk{\"u}hler, Thomas and Drettakis, George},
  journal={ACM Trans. Graph.},
  volume={42},
  number={4},
  pages={139--1},
  year={2023}
}

@article{dinov2,
  title={Dinov2: Learning robust visual features without supervision},
  author={Oquab, Maxime and Darcet, Timoth{\'e}e and Moutakanni, Th{\'e}o and Vo, Huy and Szafraniec, Marc and Khalidov, Vasil and Fernandez, Pierre and Haziza, Daniel and Massa, Francisco and El-Nouby, Alaaeldin and others},
  journal={arXiv preprint arXiv:2304.07193},
  year={2023}
}

@article{flame,
  title={Learning a model of facial shape and expression from 4D scans.},
  author={Li, Tianye and Bolkart, Timo and Black, Michael J and Li, Hao and Romero, Javier},
  journal={ACM Trans. Graph.},
  volume={36},
  number={6},
  pages={194--1},
  year={2017}
}

@article{nersemble,
    author = {Kirschstein, Tobias and Qian, Shenhan and Giebenhain, Simon and Walter, Tim and Nie\ss{}ner, Matthias},
    title = {NeRSemble: Multi-View Radiance Field Reconstruction of Human Heads},
    year = {2023},
    issue_date = {August 2023},
    publisher = {Association for Computing Machinery},
    address = {New York, NY, USA},
    volume = {42},
    number = {4},
    issn = {0730-0301},
    url = {https://doi.org/10.1145/3592455},
    doi = {10.1145/3592455},
    journal = {ACM Trans. Graph.},
    month = {jul},
    articleno = {161},
    numpages = {14},
}

@inproceedings{monogaussianavatar,
  title={Monogaussianavatar: Monocular gaussian point-based head avatar},
  author={Chen, Yufan and Wang, Lizhen and Li, Qijing and Xiao, Hongjiang and Zhang, Shengping and Yao, Hongxun and Liu, Yebin},
  booktitle={ACM SIGGRAPH 2024 Conference Papers},
  pages={1--9},
  year={2024}
}

@misc{avat3r,
      title={Avat3r: Large Animatable Gaussian Reconstruction Model for High-fidelity 3D Head Avatars},
      author={Tobias Kirschstein and Javier Romero and Artem Sevastopolsky and Matthias Nie\ss{}ner and Shunsuke Saito},
      year={2025},
      eprint={2502.20220},
      archivePrefix={arXiv},
      primaryClass={cs.CV},
      url={https://arxiv.org/abs/2502.20220},
}

@article{ava256,
  author = {Julieta Martinez and Emily Kim and Javier Romero and Timur Bagautdinov and Shunsuke Saito and Shoou-I Yu and Stuart Anderson and Michael Zollhöfer and Te-Li Wang and Shaojie Bai and Chenghui Li and Shih-En Wei and Rohan Joshi and Wyatt Borsos and Tomas Simon and Jason Saragih and Paul Theodosis and Alexander Greene and Anjani Josyula and Silvio Mano Maeta and Andrew I. Jewett and Simon Venshtain and Christopher Heilman and Yueh-Tung Chen and Sidi Fu and Mohamed Ezzeldin A. Elshaer and Tingfang Du and Longhua Wu and Shen-Chi Chen and Kai Kang and Michael Wu and Youssef Emad and Steven Longay and Ashley Brewer and Hitesh Shah and James Booth and Taylor Koska and Kayla Haidle and Matt Andromalos and Joanna Hsu and Thomas Dauer and Peter Selednik and Tim Godisart and Scott Ardisson and Matthew Cipperly and Ben Humberston and Lon Farr and Bob Hansen and Peihong Guo and Dave Braun and Steven Krenn and He Wen and Lucas Evans and Natalia Fadeeva and Matthew Stewart and Gabriel Schwartz and Divam Gupta and Gyeongsik Moon and Kaiwen Guo and Yuan Dong and Yichen Xu and Takaaki Shiratori and Fabian Prada and Bernardo R. Pires and Bo Peng and Julia Buffalini and Autumn Trimble and Kevyn McPhail and Melissa Schoeller and Yaser Sheikh},
  title = {{Codec Avatar Studio: Paired Human Captures for Complete, Driveable, and Generalizable Avatars}},
  year = {2024},
  journal = {NeurIPS Track on Datasets and Benchmarks},
}

@inproceedings{DBLP:conf/iccv/RanftlBK21,
  author       = {Ren{\'{e}} Ranftl and
                  Alexey Bochkovskiy and
                  Vladlen Koltun},
  title        = {Vision Transformers for Dense Prediction},
  booktitle    = {{ICCV} 2021},
  pages        = {12159--12168},
  publisher    = {{IEEE}},
  year         = {2021},
}

@inproceedings{DBLP:conf/cvpr/QianKS0GN24,
  author       = {Shenhan Qian and
                  Tobias Kirschstein and
                  Liam Schoneveld and
                  Davide Davoli and
                  Simon Giebenhain and
                  Matthias Nie{\ss}ner},
  title        = {GaussianAvatars: Photorealistic Head Avatars with Rigged 3D Gaussians},
  booktitle    = {{CVPR} 2024},
  pages        = {20299--20309},
  publisher    = {{IEEE}},
  year         = {2024}
}

@inproceedings{DBLP:conf/siggraph/BlanzV99,
  author       = {Volker Blanz and
                  Thomas Vetter},
  editor       = {Warren N. Waggenspack},
  title        = {A Morphable Model for the Synthesis of 3D Faces},
  booktitle    = {{SIGGRAPH} 1999},
  pages        = {187--194},
  publisher    = {{ACM}},
  year         = {1999}
}

@article{feng2021learning,
  title={Learning an animatable detailed 3D face model from in-the-wild images},
  author={Feng, Yao and Feng, Haiwen and Black, Michael J and Bolkart, Timo},
  journal={ACM Transactions on Graphics (ToG)},
  volume={40},
  number={4},
  pages={1--13},
  year={2021},
  publisher={ACM New York, NY, USA}
}

@inproceedings{danvevcek2022emoca,
  title={Emoca: Emotion driven monocular face capture and animation},
  author={Dan{\v{e}}{\v{c}}ek, Radek and Black, Michael J and Bolkart, Timo},
  booktitle={Proceedings of the IEEE/CVF Conference on Computer Vision and Pattern Recognition},
  pages={20311--20322},
  year={2022}
}

@inproceedings{ma2024cvthead,
  title={Cvthead: One-shot controllable head avatar with vertex-feature transformer},
  author={Ma, Haoyu and Zhang, Tong and Sun, Shanlin and Yan, Xiangyi and Han, Kun and Xie, Xiaohui},
  booktitle={Proceedings of the IEEE/CVF Winter Conference on Applications of Computer Vision},
  pages={6131--6141},
  year={2024}
}

@article{mildenhall2021nerf,
  title={Nerf: Representing scenes as neural radiance fields for view synthesis},
  author={Mildenhall, Ben and Srinivasan, Pratul P and Tancik, Matthew and Barron, Jonathan T and Ramamoorthi, Ravi and Ng, Ren},
  journal={Communications of the ACM},
  volume={65},
  number={1},
  pages={99--106},
  year={2021},
  publisher={ACM New York, NY, USA}
}

@article{muller2022instant,
  title={Instant neural graphics primitives with a multiresolution hash encoding},
  author={M{\"u}ller, Thomas and Evans, Alex and Schied, Christoph and Keller, Alexander},
  journal={ACM transactions on graphics (TOG)},
  volume={41},
  number={4},
  pages={1--15},
  year={2022},
  publisher={ACM New York, NY, USA}
}

@inproceedings{zielonka2023instant,
  title={Instant volumetric head avatars},
  author={Zielonka, Wojciech and Bolkart, Timo and Thies, Justus},
  booktitle={Proceedings of the IEEE/CVF conference on computer vision and pattern recognition},
  pages={4574--4584},
  year={2023}
}

@inproceedings{shao2023tensor4d,
  title={Tensor4d: Efficient neural 4d decomposition for high-fidelity dynamic reconstruction and rendering},
  author={Shao, Ruizhi and Zheng, Zerong and Tu, Hanzhang and Liu, Boning and Zhang, Hongwen and Liu, Yebin},
  booktitle={Proceedings of the IEEE/CVF Conference on Computer Vision and Pattern Recognition},
  pages={16632--16642},
  year={2023}
}

@inproceedings{athar2023flame,
  title={Flame-in-nerf: Neural control of radiance fields for free view face animation},
  author={Athar, ShahRukh and Shu, Zhixin and Samaras, Dimitris},
  booktitle={2023 IEEE 17th International Conference on Automatic Face and Gesture Recognition (FG)},
  pages={1--8},
  year={2023},
  organization={IEEE}
}

@inproceedings{cudeiro2019capture,
  title={Capture, learning, and synthesis of 3D speaking styles},
  author={Cudeiro, Daniel and Bolkart, Timo and Laidlaw, Cassidy and Ranjan, Anurag and Black, Michael J},
  booktitle={Proceedings of the IEEE/CVF conference on computer vision and pattern recognition},
  pages={10101--10111},
  year={2019}
}

@inproceedings{athar2022rignerf,
  title={Rignerf: Fully controllable neural 3d portraits},
  author={Athar, ShahRukh and Xu, Zexiang and Sunkavalli, Kalyan and Shechtman, Eli and Shu, Zhixin},
  booktitle={Proceedings of the IEEE/CVF conference on Computer Vision and Pattern Recognition},
  pages={20364--20373},
  year={2022}
}

@inproceedings{guo2021ad,
  title={Ad-nerf: Audio driven neural radiance fields for talking head synthesis},
  author={Guo, Yudong and Chen, Keyu and Liang, Sen and Liu, Yong-Jin and Bao, Hujun and Zhang, Juyong},
  booktitle={Proceedings of the IEEE/CVF international conference on computer vision},
  pages={5784--5794},
  year={2021}
}

@inproceedings{liu2022semantic,
  title={Semantic-aware implicit neural audio-driven video portrait generation},
  author={Liu, Xian and Xu, Yinghao and Wu, Qianyi and Zhou, Hang and Wu, Wayne and Zhou, Bolei},
  booktitle={European conference on computer vision},
  pages={106--125},
  year={2022},
  organization={Springer}
}

@inproceedings{xu2024gaussian,
  title={Gaussian head avatar: Ultra high-fidelity head avatar via dynamic gaussians},
  author={Xu, Yuelang and Chen, Benwang and Li, Zhe and Zhang, Hongwen and Wang, Lizhen and Zheng, Zerong and Liu, Yebin},
  booktitle={Proceedings of the IEEE/CVF conference on computer vision and pattern recognition},
  pages={1931--1941},
  year={2024}
}

@article{wang2025gaussianhead,
  title={Gaussianhead: High-fidelity head avatars with learnable gaussian derivation},
  author={Wang, Jie and Xie, Jiu-Cheng and Li, Xianyan and Xu, Feng and Pun, Chi-Man and Gao, Hao},
  journal={IEEE Transactions on Visualization and Computer Graphics},
  year={2025},
  publisher={IEEE}
}

@inproceedings{schonberger2016structure,
  title={Structure-from-motion revisited},
  author={Schonberger, Johannes L and Frahm, Jan-Michael},
  booktitle={Proceedings of the IEEE conference on computer vision and pattern recognition},
  pages={4104--4113},
  year={2016}
}

@inproceedings{schonberger2016pixelwise,
  title={Pixelwise view selection for unstructured multi-view stereo},
  author={Sch{\"o}nberger, Johannes L and Zheng, Enliang and Frahm, Jan-Michael and Pollefeys, Marc},
  booktitle={European conference on computer vision},
  pages={501--518},
  year={2016},
  organization={Springer}
}

@inproceedings{wang2024dust3r,
  title={Dust3r: Geometric 3d vision made easy},
  author={Wang, Shuzhe and Leroy, Vincent and Cabon, Yohann and Chidlovskii, Boris and Revaud, Jerome},
  booktitle={Proceedings of the IEEE/CVF Conference on Computer Vision and Pattern Recognition},
  pages={20697--20709},
  year={2024}
}

@inproceedings{chen2024mvsplat,
  title={Mvsplat: Efficient 3d gaussian splatting from sparse multi-view images},
  author={Chen, Yuedong and Xu, Haofei and Zheng, Chuanxia and Zhuang, Bohan and Pollefeys, Marc and Geiger, Andreas and Cham, Tat-Jen and Cai, Jianfei},
  booktitle={European Conference on Computer Vision},
  pages={370--386},
  year={2024},
  organization={Springer}
}

@article{liu2024meshformer,
  title={Meshformer: High-quality mesh generation with 3d-guided reconstruction model},
  author={Liu, Minghua and Zeng, Chong and Wei, Xinyue and Shi, Ruoxi and Chen, Linghao and Xu, Chao and Zhang, Mengqi and Wang, Zhaoning and Zhang, Xiaoshuai and Liu, Isabella and others},
  journal={Advances in Neural Information Processing Systems},
  volume={37},
  pages={59314--59341},
  year={2024}
}

@article{ye2024no,
  title={No pose, no problem: Surprisingly simple 3d gaussian splats from sparse unposed images},
  author={Ye, Botao and Liu, Sifei and Xu, Haofei and Li, Xueting and Pollefeys, Marc and Yang, Ming-Hsuan and Peng, Songyou},
  journal={arXiv preprint arXiv:2410.24207},
  year={2024}
}

@article{hong2023lrm,
  title={Lrm: Large reconstruction model for single image to 3d},
  author={Hong, Yicong and Zhang, Kai and Gu, Jiuxiang and Bi, Sai and Zhou, Yang and Liu, Difan and Liu, Feng and Sunkavalli, Kalyan and Bui, Trung and Tan, Hao},
  journal={arXiv preprint arXiv:2311.04400},
  year={2023}
}

@inproceedings{charatan2024pixelsplat,
  title={pixelsplat: 3d gaussian splats from image pairs for scalable generalizable 3d reconstruction},
  author={Charatan, David and Li, Sizhe Lester and Tagliasacchi, Andrea and Sitzmann, Vincent},
  booktitle={Proceedings of the IEEE/CVF conference on computer vision and pattern recognition},
  pages={19457--19467},
  year={2024}
}

@inproceedings{szymanowicz2024splatter,
  title={Splatter image: Ultra-fast single-view 3d reconstruction},
  author={Szymanowicz, Stanislaw and Rupprecht, Chrisitian and Vedaldi, Andrea},
  booktitle={Proceedings of the IEEE/CVF conference on computer vision and pattern recognition},
  pages={10208--10217},
  year={2024}
}

@inproceedings{tang2024lgm,
  title={Lgm: Large multi-view gaussian model for high-resolution 3d content creation},
  author={Tang, Jiaxiang and Chen, Zhaoxi and Chen, Xiaokang and Wang, Tengfei and Zeng, Gang and Liu, Ziwei},
  booktitle={European Conference on Computer Vision},
  pages={1--18},
  year={2024},
  organization={Springer}
}

@article{jin2024lvsm,
  title={Lvsm: A large view synthesis model with minimal 3d inductive bias},
  author={Jin, Haian and Jiang, Hanwen and Tan, Hao and Zhang, Kai and Bi, Sai and Zhang, Tianyuan and Luan, Fujun and Snavely, Noah and Xu, Zexiang},
  journal={arXiv preprint arXiv:2410.17242},
  year={2024}
}

@article{zhang2024monst3r,
  title={Monst3r: A simple approach for estimating geometry in the presence of motion},
  author={Zhang, Junyi and Herrmann, Charles and Hur, Junhwa and Jampani, Varun and Darrell, Trevor and Cole, Forrester and Sun, Deqing and Yang, Ming-Hsuan},
  journal={arXiv preprint arXiv:2410.03825},
  year={2024}
}

@misc{labs2025flux1kontextflowmatching,
      title={FLUX.1 Kontext: Flow Matching for In-Context Image Generation and Editing in Latent Space},
      author={Black Forest Labs and Stephen Batifol and Andreas Blattmann and Frederic Boesel and Saksham Consul and Cyril Diagne and Tim Dockhorn and Jack English and Zion English and Patrick Esser and Sumith Kulal and Kyle Lacey and Yam Levi and Cheng Li and Dominik Lorenz and Jonas Müller and Dustin Podell and Robin Rombach and Harry Saini and Axel Sauer and Luke Smith},
      year={2025},
      eprint={2506.15742},
      archivePrefix={arXiv},
      primaryClass={cs.GR},
      url={https://arxiv.org/abs/2506.15742},
}

@misc{flux2024,
    author={Black Forest Labs},
    title={FLUX},
    year={2024},
    howpublished={\url{https://github.com/black-forest-labs/flux}},
}

@inproceedings{DBLP:conf/iclr/ChuLZYLLH24,
  author       = {Xuangeng Chu and
                  Yu Li and
                  Ailing Zeng and
                  Tianyu Yang and
                  Lijian Lin and
                  Yunfei Liu and
                  Tatsuya Harada},
  title        = {GPAvatar: Generalizable and Precise Head Avatar from Image(s)},
  booktitle    = {{ICLR} 2024},
  publisher    = {OpenReview.net},
  year         = {2024}
}

@inproceedings{DBLP:conf/iclr/JangGP17,
  author       = {Eric Jang and
                  Shixiang Gu and
                  Ben Poole},
  title        = {Categorical Reparameterization with Gumbel-Softmax},
  booktitle    = {5th International Conference on Learning Representations, {ICLR} 2017,
                  Toulon, France, April 24-26, 2017, Conference Track Proceedings},
  publisher    = {OpenReview.net},
  year         = {2017},
  url          = {https://openreview.net/forum?id=rkE3y85ee},
  bibsource    = {dblp computer science bibliography, https://dblp.org}
}

@article{DBLP:journals/corr/abs-2505-23716,
  author       = {Lihan Jiang and
                  Yucheng Mao and
                  Linning Xu and
                  Tao Lu and
                  Kerui Ren and
                  Yichen Jin and
                  Xudong Xu and
                  Mulin Yu and
                  Jiangmiao Pang and
                  Feng Zhao and
                  Dahua Lin and
                  Bo Dai},
  title        = {AnySplat: Feed-forward 3D Gaussian Splatting from Unconstrained Views},
  journal      = {CoRR},
  volume       = {abs/2505.23716},
  year         = {2025},
  url          = {https://doi.org/10.48550/arXiv.2505.23716},
  doi          = {10.48550/ARXIV.2505.23716},
  eprinttype    = {arXiv},
  eprint       = {2505.23716}
}

@inproceedings{DBLP:conf/nips/Pan0L0ZLSML24,
  author       = {Panwang Pan and
                  Zhuo Su and
                  Chenguo Lin and
                  Zhen Fan and
                  Yongjie Zhang and
                  Zeming Li and
                  Tingting Shen and
                  Yadong Mu and
                  Yebin Liu},
  title        = {HumanSplat: Generalizable Single-Image Human Gaussian Splatting with
                  Structure Priors},
  booktitle    = {Advances in Neural Information Processing Systems 38: Annual Conference
                  on Neural Information Processing Systems 2024, NeurIPS 2024, Vancouver,
                  BC, Canada, December 10 - 15, 2024},
  year         = {2024}
}

@inproceedings{DBLP:conf/cvpr/Wen0RSW24,
  author       = {Jing Wen and
                  Xiaoming Zhao and
                  Zhongzheng Ren and
                  Alexander G. Schwing and
                  Shenlong Wang},
  title        = {GoMAvatar: Efficient Animatable Human Modeling from Monocular Video
                  Using Gaussians-on-Mesh},
  booktitle    = {{IEEE/CVF} Conference on Computer Vision and Pattern Recognition,
                  {CVPR} 2024, Seattle, WA, USA, June 16-22, 2024},
  pages        = {2059--2069},
  publisher    = {{IEEE}},
  year         = {2024}
}

@inproceedings{DBLP:conf/cvpr/QianWM0024,
  author       = {Zhiyin Qian and
                  Shaofei Wang and
                  Marko Mihajlovic and
                  Andreas Geiger and
                  Siyu Tang},
  title        = {3DGS-Avatar: Animatable Avatars via Deformable 3D Gaussian Splatting},
  booktitle    = {{IEEE/CVF} Conference on Computer Vision and Pattern Recognition,
                  {CVPR} 2024, Seattle, WA, USA, June 16-22, 2024},
  pages        = {5020--5030},
  publisher    = {{IEEE}},
  year         = {2024}
}

@inproceedings{DBLP:conf/cvpr/JiangSW0HZYX24,
  author       = {Yuheng Jiang and
                  Zhehao Shen and
                  Penghao Wang and
                  Zhuo Su and
                  Yu Hong and
                  Yingliang Zhang and
                  Jingyi Yu and
                  Lan Xu},
  title        = {HiFi4G: High-Fidelity Human Performance Rendering via Compact Gaussian
                  Splatting},
  booktitle    = {{IEEE/CVF} Conference on Computer Vision and Pattern Recognition,
                  {CVPR} 2024, Seattle, WA, USA, June 16-22, 2024},
  pages        = {19734--19745},
  publisher    = {{IEEE}},
  year         = {2024}
}

@inproceedings{DBLP:conf/cvpr/Hu00ZLZN24,
  author       = {Liangxiao Hu and
                  Hongwen Zhang and
                  Yuxiang Zhang and
                  Boyao Zhou and
                  Boning Liu and
                  Shengping Zhang and
                  Liqiang Nie},
  title        = {GaussianAvatar: Towards Realistic Human Avatar Modeling from a Single
                  Video via Animatable 3D Gaussians},
  booktitle    = {{IEEE/CVF} Conference on Computer Vision and Pattern Recognition,
                  {CVPR} 2024, Seattle, WA, USA, June 16-22, 2024},
  pages        = {634--644},
  publisher    = {{IEEE}},
  year         = {2024}
}

@article{DBLP:journals/corr/abs-2503-10625,
  author       = {Lingteng Qiu and
                  Xiaodong Gu and
                  Peihao Li and
                  Qi Zuo and
                  Weichao Shen and
                  Junfei Zhang and
                  Kejie Qiu and
                  Weihao Yuan and
                  Guanying Chen and
                  Zilong Dong and
                  Liefeng Bo},
  title        = {{LHM:} Large Animatable Human Reconstruction Model from a Single Image
                  in Seconds},
  journal      = {CoRR},
  volume       = {abs/2503.10625},
  year         = {2025}
}

@inproceedings{LP3DGS,
 author = {Zhang, Zhaoliang and Song, Tianchen and Lee, Yongjae and Yang, Li and Peng, Cheng and Chellappa, Rama and Fan, Deliang},
 booktitle = {Advances in Neural Information Processing Systems},
 editor = {A. Globerson and L. Mackey and D. Belgrave and A. Fan and U. Paquet and J. Tomczak and C. Zhang},
 pages = {122434--122457},
 publisher = {Curran Associates, Inc.},
 title = {LP-3DGS: Learning to Prune 3D Gaussian Splatting},
 url = {https://proceedings.neurips.cc/paper_files/paper/2024/file/dd51dbce305433cd60910dc5b0147be4-Paper-Conference.pdf},
 volume = {37},
 year = {2024}
}

@InProceedings{MaskGaussian,
    author    = {Liu, Yifei and Zhong, Zhihang and Zhan, Yifan and Xu, Sheng and Sun, Xiao},
    title     = {MaskGaussian: Adaptive 3D Gaussian Representation from Probabilistic Masks},
    booktitle = {Proceedings of the Computer Vision and Pattern Recognition Conference (CVPR)},
    month     = {June},
    year      = {2025},
    pages     = {681-690}
}

@inproceedings{zhang2025framepack,
    title={Frame Context Packing and Drift Prevention in Next-Frame-Prediction Video Diffusion Models},
    author={Lvmin Zhang and Shengqu Cai and Muyang Li and Gordon Wetzstein and Maneesh Agrawala},
    booktitle={The Thirty-ninth Annual Conference on Neural Information Processing Systems},
    year={2025},
}

@inproceedings{gerogiannis2025arc2avatar,
  title={Arc2avatar: Generating expressive 3d avatars from a single image via id guidance},
  author={Gerogiannis, Dimitrios and Papantoniou, Foivos Paraperas and Potamias, Rolandos Alexandros and Lattas, Alexandros and Zafeiriou, Stefanos},
  booktitle={Proceedings of the Computer Vision and Pattern Recognition Conference},
  pages={10770--10782},
  year={2025}
}

@inproceedings{zielonka2025synthetic,
  title={Synthetic prior for few-shot drivable head avatar inversion},
  author={Zielonka, Wojciech and Garbin, Stephan J and Lattas, Alexandros and Kopanas, George and Gotardo, Paulo and Beeler, Thabo and Thies, Justus and Bolkart, Timo},
  booktitle={Proceedings of the Computer Vision and Pattern Recognition Conference},
  pages={10735--10746},
  year={2025}
}

@inproceedings{zheng2025headgap,
  title={Headgap: Few-shot 3d head avatar via generalizable gaussian priors},
  author={Zheng, Xiaozheng and Wen, Chao and Li, Zhaohu and Zhang, Weiyi and Su, Zhuo and Chang, Xu and Zhao, Yang and Lv, Zheng and Zhang, Xiaoyuan and Zhang, Yongjie and others},
  booktitle={2025 International Conference on 3D Vision (3DV)},
  pages={946--957},
  year={2025},
  organization={IEEE}
}

@inproceedings{deng2019arcface,
  title={Arcface: Additive angular margin loss for deep face recognition},
  author={Deng, Jiankang and Guo, Jia and Xue, Niannan and Zafeiriou, Stefanos},
  booktitle={Proceedings of the IEEE/CVF conference on computer vision and pattern recognition},
  pages={4690--4699},
  year={2019}
}

@inproceedings{DBLP:conf/mm/ChenCNG20,
  author       = {Renwang Chen and
                  Xuanhong Chen and
                  Bingbing Ni and
                  Yanhao Ge},
  editor       = {Chang Wen Chen and
                  Rita Cucchiara and
                  Xian{-}Sheng Hua and
                  Guo{-}Jun Qi and
                  Elisa Ricci and
                  Zhengyou Zhang and
                  Roger Zimmermann},
  title        = {SimSwap: An Efficient Framework For High Fidelity Face Swapping},
  booktitle    = {{MM} '20: The 28th {ACM} International Conference on Multimedia, Virtual
                  Event / Seattle, WA, USA, October 12-16, 2020},
  pages        = {2003--2011},
  publisher    = {{ACM}},
  year         = {2020},
  doi          = {10.1145/3394171.3413630},
  bibsource    = {dblp computer science bibliography, https://dblp.org}
}

@article{DBLP:journals/pami/ChenNLLZW24,
  author       = {Xuanhong Chen and
                  Bingbing Ni and
                  Yutian Liu and
                  Naiyuan Liu and
                  Zhilin Zeng and
                  Hang Wang},
  title        = {SimSwap++: Towards Faster and High-Quality Identity Swapping},
  journal      = {{IEEE} Trans. Pattern Anal. Mach. Intell.},
  volume       = {46},
  number       = {1},
  pages        = {576--592},
  year         = {2024},
  doi          = {10.1109/TPAMI.2023.3307156},
  bibsource    = {dblp computer science bibliography, https://dblp.org}
}

@inproceedings{DBLP:conf/cvpr/WuCLJWFCLHZNZ25,
  author       = {Yufan Wu and
                  Xuanhong Chen and
                  Wen Li and
                  Shunran Jia and
                  Hualiang Wei and
                  Kairui Feng and
                  Jialiang Chen and
                  Yuhan Li and
                  Ang He and
                  Weimin Zhang and
                  Bingbing Ni and
                  Wenjun Zhang},
  title        = {SinGS: Animatable Single-Image Human Gaussian Splats with Kinematic
                  Priors},
  booktitle    = {{IEEE/CVF} Conference on Computer Vision and Pattern Recognition,
                  {CVPR} 2025, Nashville, TN, USA, June 11-15, 2025},
  pages        = {5571--5580},
  publisher    = {Computer Vision Foundation / {IEEE}},
  year         = {2025},
  doi          = {10.1109/CVPR52734.2025.00523},
  bibsource    = {dblp computer science bibliography, https://dblp.org}
}
\bibliographystyle{iclr2026_conference}

\appendix
\newpage
\section{LLM Usage}
Large Language Models (LLMs) were used solely to assist in refining the manuscript’s language, improving readability, and ensuring clarity, including sentence rephrasing and grammar checking. The LLM did not contribute to the research ideas, methodology, experiments, or data analysis. The authors take full responsibility for the scientific content and confirm that all LLM-assisted text adheres to ethical guidelines, with no contribution to plagiarism or misconduct.

\section{Ethics Statement}
The proposed FastAvatar framework follows the same data assumptions and usage boundaries as existing 3D avatar reconstruction and neural rendering methods. It does not introduce new mechanisms that lower the barrier to unauthorized identity reconstruction, nor does it relax requirements on input data quality. In practice, the method still relies on clean and identity-consistent multi-frame input, which inherently limits large-scale or covert misuse.

All experiments are conducted on publicly available datasets with appropriate licenses, and no private or sensitive data are collected. The intended applications of FastAvatar lie in areas such as AR/VR telepresence, digital content creation, and human–computer interaction, where user consent is typically explicit. We emphasize that responsible deployment requires ensuring that the method is not applied to reconstruct individuals without consent or to generate deceptive or impersonating content.

\section{reproducibility}
In this section, we provide more implementation details of FastAvatar, including data preparation and model architecture. Furthermore, our code will be released after the paper is accepted. 

\subsection{Data Preparation}
Our training utilizes the Nersemble dataset. Initially, FLAME tracking is applied to extract FLAME parameters and camera poses, which serve as inputs for subsequent training stages. From the original Nersemble data, we extract 521 distinct video clips, and sample the frames at 15 FPS. Cameras with poor face tracking quality were excluded, remaining 12 cameras. The processed data was sampled twice to construct the final dataset: first, sampling frames within the same video sequence, and second, performing random sampling across all shots of the same action sequence. These two sampling strategies collectively support training the unified task. To enhance the stability of training and testing, we randomly assign the processed figures’ backgrounds to black, white, or gray. Note that, to enhance the model's generative capability, all expression parameters, poses, and related data used during inference differ from the input data.

\begin{table}[htb]
    \centering
    \begin{tabular}{lll}
        \toprule
        & Hyperparameter & Value \\
        \midrule
            \multirowcell{7}{\rotatebox[origin=c]{90}{{\centering \small Encoder}}} 
                & DINOv2 patch size & $14 \times 14$ \\
                & \#expression token MLP layers & 2 \\
                & \#camera token MLP layers & 2 \\
                & Expression Token MLP activation & GELU \\
                & Camera Token MLP activation & GELU \\
                & Output dimension & 1024 \\
                & Input image resolution & $504 \times 504$ \\
         \midrule
            \multirowcell{4}{\rotatebox[origin=c]{90}{{\centering \small AA}}} 
                & \#Frame Attn Layers & 10 \\
                & \#Global Attn Layers & 10 \\
                & Hidden dimension & 1024  \\
                & Order & [Global, Frame] \\
         \bottomrule
         
    \end{tabular}
    \caption{Hyperparameters. Where AA represents Alternation Attention}
    \label{tab:hyperparameters}
\end{table}

The accuracy of FLAME tracking primarily depends on the precision of detected facial landmarks, as the FLAME parameters are typically estimated by optimizing the model to fit these landmarks. However, the reliability of such proxy models (e.g., FLAME and other 3DMMs) is inherently constrained by factors such as limited representational capacity and sensitivity to landmark quality. To assess how these factors affect our method, we include an additional ablation experiment that injects controlled noise into the facial landmarks. The results (Table~\ref{tab:ablation-flame}) demonstrate that FastAvatar remains robust under reasonable perturbations, indicating that strict accuracy in FLAME tracking is not required.

\begin{table*}[tp]
\centering
    \begin{threeparttable}[b]
    \resizebox{0.48\textwidth}{!}{
    \begin{tabular}{*{10}{l|ccccc}}
        \toprule
         Noise &  L1$\downarrow$ & PSNR$\uparrow$ &  SSIM$\uparrow$ & LPIPS$\downarrow$ & Identity$\downarrow$ \\
        \midrule
        1 px & \underline{0.0264} & \textbf{22.50} & \textbf{0.873} & \textbf{0.096} & \underline{0.100} \\
        4 px & 0.0268 & \underline{22.38} & 0.872 & \underline{0.098} & \textbf{0.099} \\
        8 px & 0.0273 & 22.22 & 0.870 & \underline{0.098} & 0.103 \\
        16 px & 0.0277 & 22.10 & 0.869 & 0.099 & 0.102 \\
        32 px & 0.0280 & 22.02 & 0.869 & 0.099 & 0.105 \\
        Ours & \textbf{0.0263} & \textbf{22.50} & \underline{0.872} & \textbf{0.096} & \textbf{0.099} \\
        \bottomrule
    \end{tabular}
    }
  \end{threeparttable}
  \caption{Ablation studies on FLAME tracking. We evaluate the robustness of FastAvatar under varying levels of landmark perturbation during FLAME tracking.}
\label{tab:ablation-flame}
\end{table*}

\subsection{Training}
In table~\ref{tab:hyperparameters}, we present the most important hyperparameters for training FastAvatar. We train the model by optimizing the training loss with the AdamW optimizer for 150K iterations. We use a cosine learning rate scheduler with learning rate of 4e-5. The training runs on 8 H100 GPUs over 14 days. The substantial accumulation of Gaussian points across multiple input frames leads to high GPU memory consumption during training. To address this, we adopt bfloat16 precision and gradient checkpointing for improved memory and computational efficiency.

\subsection{FramePack}
While 16 sparse frames suffice for high-quality 3D head reconstruction, quality degrades when viewpoint coverage is incomplete—e.g., mouth-opening expressions fail without intra-oral views. However, full global attention over all frames is prohibitively expensive.

We address this with a two-tier token design inspired by FramePack. We designate 16 base frames whose DINO features are kept at full spatial resolution, and compress all remaining frames with a learned $k{\times}k$ strided convolution ($k{=}8$), reducing each to $1/k^2$ of its original token count. Within each alternating-attention layer, frame attention cross-attends each frame's point cloud with its own image tokens independently—base frames at full resolution, compressed frames through a separate set of weights at reduced resolution. Global attention then concatenates all tokens across both tiers and applies RoPE-based self-attention jointly. This provides coverage from arbitrarily many extra viewpoints at a cost sublinear in total frame count, enabling incremental reconstruction from hundreds of input frames.

\section{More results}
In this section, we present additional results of FastAvatar, including its performance on a broader set of video sequences and its generalization to real-world daily-captured data.

\begin{figure*}[htbp!]
  \centering
  \includegraphics[width=1.0\linewidth]{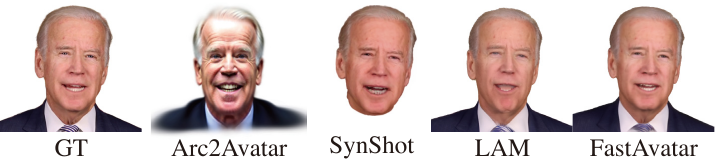}
  \caption{Qualitative results on the INSTA dataset. LPIPS scores: Ours (\textbf{0.1332}), LAM (0.1479), Arc2Avatar (0.4585), SynShot (0.1523). Identity scores: Ours (\textbf{0.076}), LAM (0.124), Arc2Avatar (0.411), SynShot (0.115).}
  \label{fig:biden}
\end{figure*}

\begin{figure*}[htbp!]
  \centering
  \includegraphics[width=1.0\linewidth]{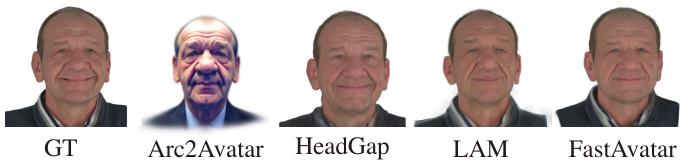}
  \caption{Qualitative results on the Nersemble dataset. LPIPS scores: Ours (\textbf{0.1267}), LAM (0.1608), Arc2Avatar (0.4665), HeadGap (0.1592). Identity scores: Ours (\textbf{0.070}), LAM (0.097), Arc2Avatar (0.308), HeadGap (0.101).}
  \label{fig:210}
\end{figure*}

\begin{figure*}[htbp]
  \centering
  \includegraphics[width=1.0\linewidth]{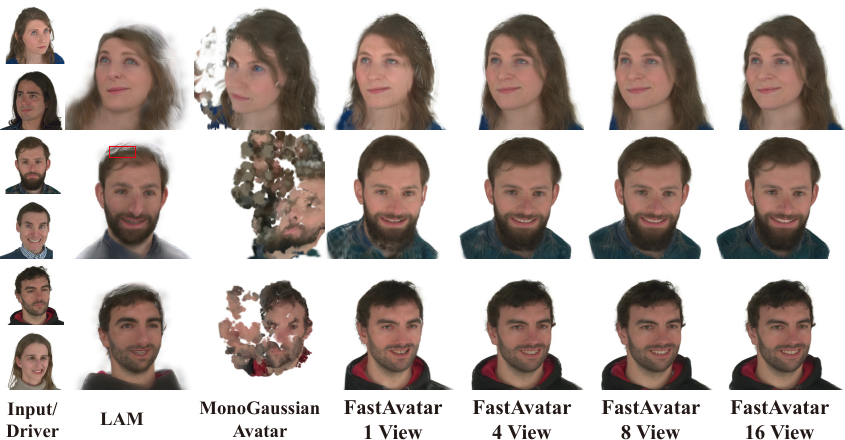}
  \caption{Additional Comparisons with Baseline Methods (cross-reenactment).}
  \label{fig:main_comp3}
\end{figure*}

\begin{figure*}[tbp]
  \centering
  \includegraphics[width=1.0\linewidth]{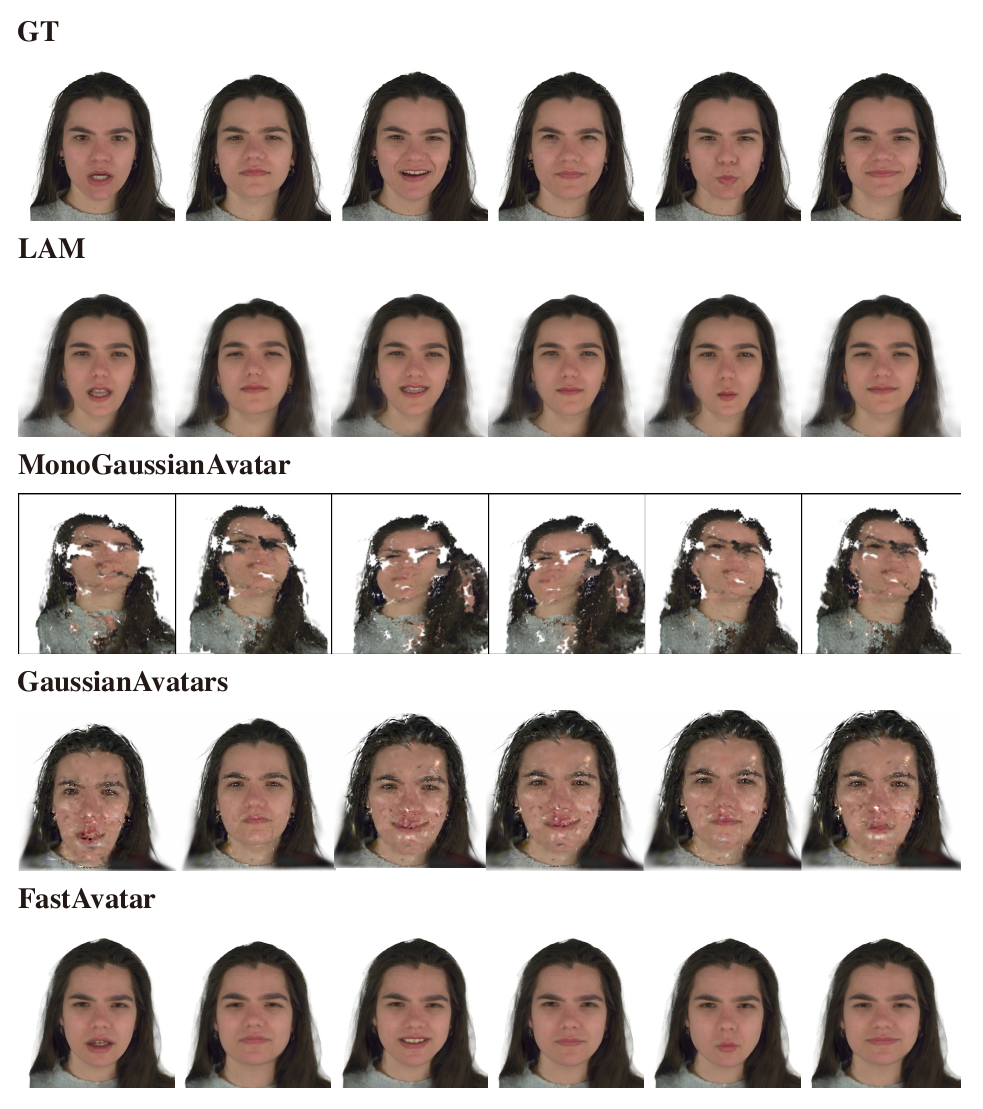}
  \caption{Additional Comparisons with Baseline Methods (self-reenactment part 1).}
  \label{fig:main_comp1}
\end{figure*}

\begin{figure*}[tbp]
  \centering
  \includegraphics[width=1.0\linewidth]{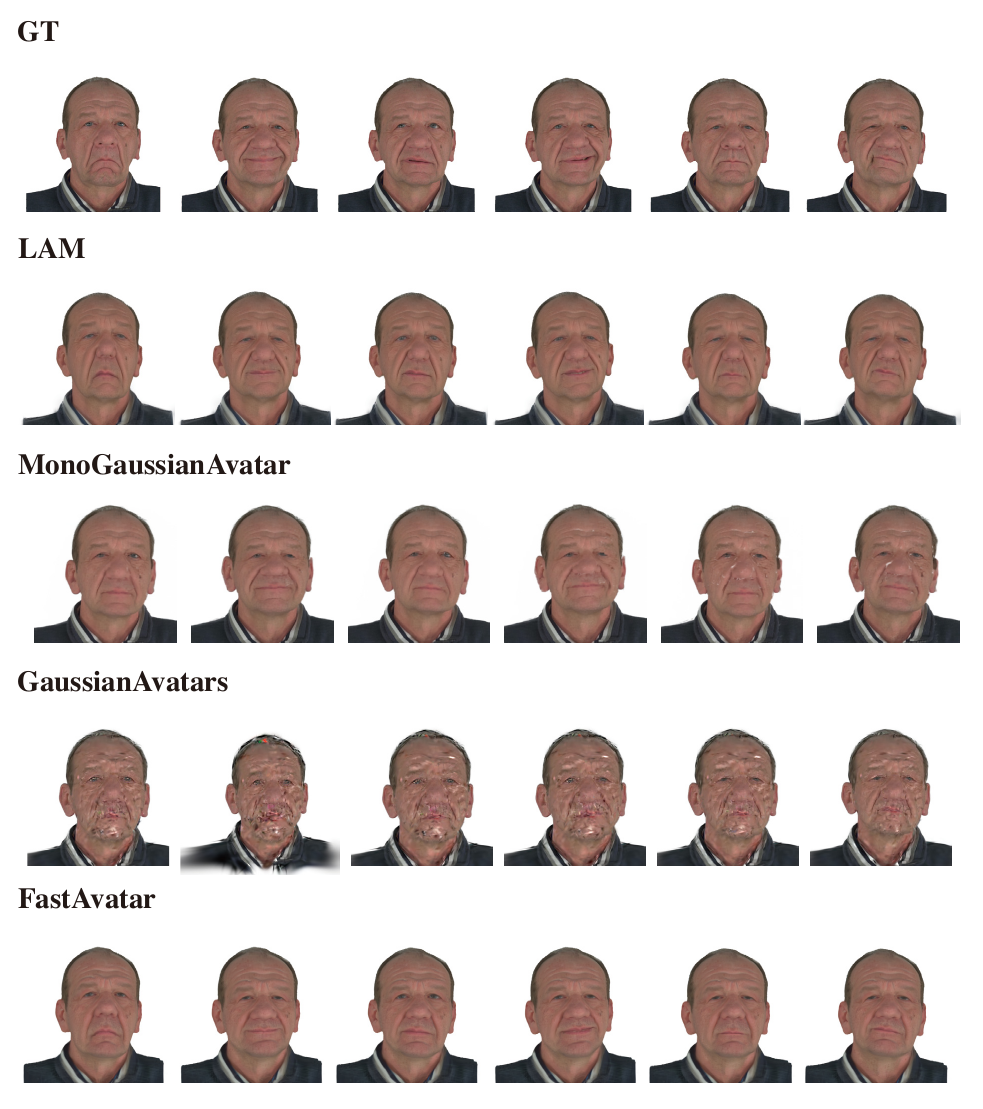}
  \caption{Additional Comparisons with Baseline Methods (self-reenactment part 2).}
  \label{fig:main_comp2}
\end{figure*}

\paragraph{More Comparison} We provide additional qualitative results comparing FastAvatar and baseline models in both self-reenactment and cross-reenactment settings. As shown in Figure~\ref{fig:main_comp1}, Figure~\ref{fig:main_comp2}, and Figure~\ref{fig:main_comp3}. FastAvatar outperforms the baselines. Optimization-based 3D avatar methods fail to achieve satisfactory results with sparse inputs, while LAM often exhibits unrealistic details and significant pose inaccuracies. The advantage of FastAvatar becomes even more evident in the cross-reenactment setting, where the subject’s identity and camera pose exhibit large discrepancies. We further evaluate FastAvatar against additional competitive methods. The results are presented in Figure~\ref{fig:biden} and Figure~\ref{fig:210}.

\begin{figure*}
    \centering
    \includegraphics[width=1.0\linewidth]{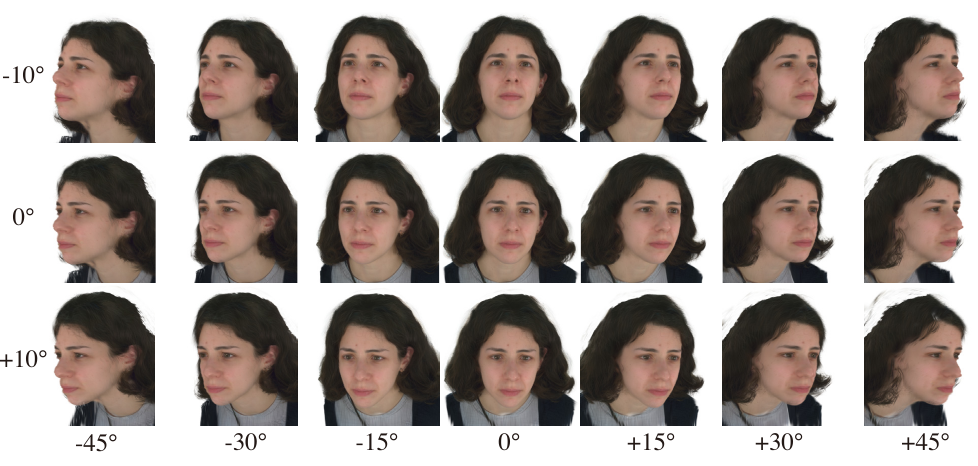}
    \caption{Generalization to wide-range viewpoints. FastAvatar achieves high-fidelity reconstruction across 14 novel viewpoints that are entirely outside the training set.}
    \label{fig:widerange}
\end{figure*}

\begin{figure*}
    \centering
    \includegraphics[width=1.0\linewidth]{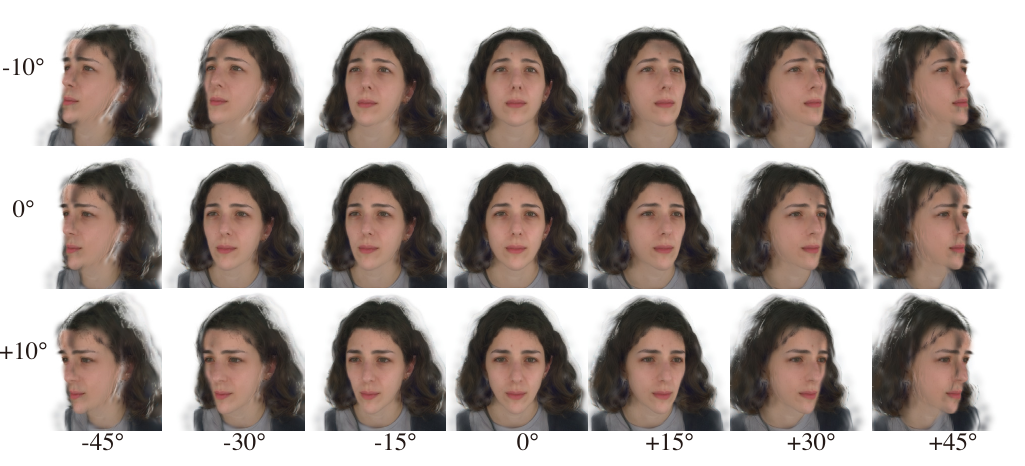}
    \caption{The performance of LAM on wide-range viewpoints.}
    \label{fig:widerange_lam}
\end{figure*}

\paragraph{Generalization to Wide-Range Viewpoints}
The Nersemble training set contains only 12 constrained camera views. To evaluate the robustness of our method, we test it on a much wider range of viewpoints. As shown in Figure~\ref{fig:widerange}, the model maintains high-fidelity reconstruction across all novel views, demonstrating strong wide-range generalization. For comparison, we include the results of LAM in Figure~\ref{fig:widerange_lam}. The results demonstrate that FastAvatar outperforms the state-of-the-art across a wide range of viewpoints.

\begin{figure*}[tbp]
  \centering
  \includegraphics[width=1.0\linewidth]{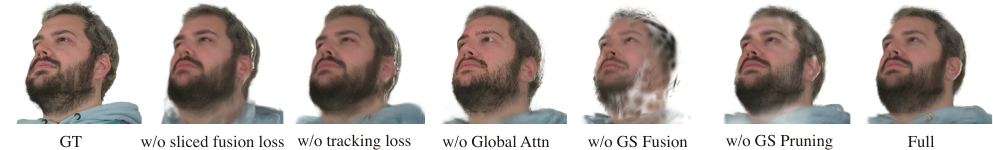}
  \caption{Comparison of visual effects of model reconstruction after removing the key components.}
  \label{fig:ablation}
\end{figure*}

\paragraph{Ablation Study}
We further highlight the role of the key components in incremental reconstruction. As illustrated in Figure~\ref{fig:ablation}, removing these components prevents fine details from being properly aligned, leading to noticeable artifacts and blurred regions. Although the landmark tracking loss only supervises 68 facial landmark points, it still provides strong guidance for Gaussian registration, effectively assisting the model in aligning new frames during incremental updates. Together with the Sliced Fusion Loss, it ensures that additional observations can be accurately fused, enabling consistent refinement of the reconstructed avatar. Global Attention enables the model to leverage inter-frame dependencies, integrating complementary features from multiple frames; without it, information remains localized, and cross-frame consistency cannot be achieved. GS Fusion consolidates per-frame Gaussians into a coherent representation, allowing the model to maintain consistency across frames. Finally, Gaussian Pruning removes redundant primitives, slightly improving performance while significantly accelerating rendering, enabling efficient incremental updates even for long sequences.

\begin{figure*}[tbp]
  \centering
  \includegraphics[width=1.0\linewidth]{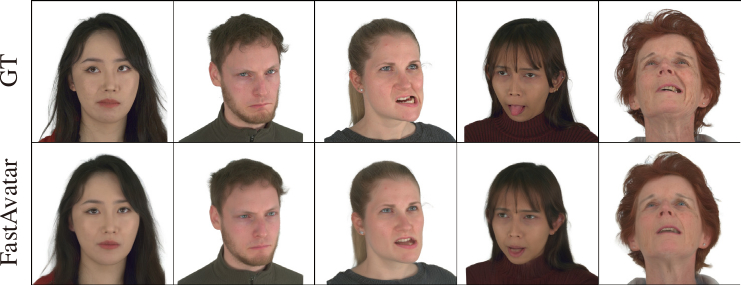}
  \caption{Typical failure cases: FastAvatar relying on LBS and FLAME-based encodings, struggles with complex facial muscle dynamics, fine-grained details (e.g., wrinkles), eye-gaze movements, and structures outside the FLAME topology such as the tongue.}
  \label{fig:limitation}
\end{figure*}

\begin{figure*}[tbp]
  \centering
  \includegraphics[width=1.0\linewidth]{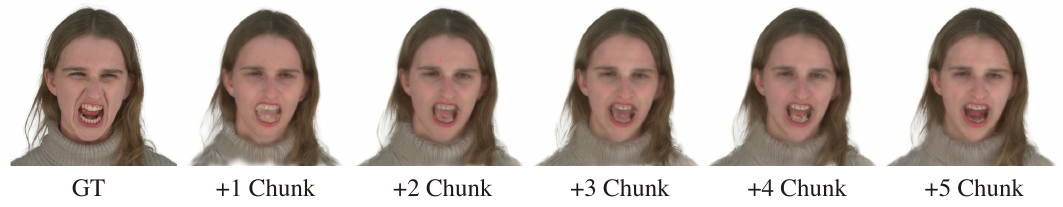}
  \caption{As the streaming input is progressively incorporated, the reconstruction of the oral cavity—largely invisible in most chunks—gradually improves while structural consistency is maintained in other regions, enabling incremental reconstruction.}
  \label{fig:steaming}
\end{figure*}

\paragraph{Streaming Incremental Reconstruction} FastAvatar is also capable of streaming incremental reconstruction, meaning the model continuously updates and refines the 3DGS representation as new video frames are received. To achieve this, we adopt a sliding-window approach, where each window contains 16 frames with a 4-frame overlap for registering incoming frames. Thanks to the Alternating Attention design, new frames can build upon the Gaussians predicted by the previous model to produce more informative reconstructions. Figure\ref{fig:steaming} demonstrates this: starting from a single-view image, as additional views are provided (excluding the test view), reconstruction quality improves overall, including at previously unseen angles, thus realizing streaming incremental reconstruction.

\paragraph{Limitations}
First, our method relies on LBS and FLAME-based encodings to drive 3D head avatar motion, which limits the representation of complex facial muscle dynamics. As a result, the model has difficulty reproducing fine-grained muscle-dependent details such as wrinkles and also cannot accurately capture eye-gaze movements, often defaulting to an average direction. Furthermore, because the Gaussians are anchored to FLAME vertices, the model is unable to represent structures outside the FLAME topology, including the tongue. Figure~\ref{fig:limitation} presents representative failure cases.

\end{document}